\documentclass[letterpaper, 10 pt, conference]{ieeeconf}  

\IEEEoverridecommandlockouts                              

\usepackage{graphics} 
\usepackage{epsfig} 
\usepackage{times} 
\usepackage{amsmath} 
\usepackage{amssymb}  
\usepackage{cite}
\usepackage{subcaption}
\usepackage[normalem]{ulem}
\usepackage[ruled,vlined,linesnumbered]{algorithm2e}
\usepackage{xargs}    
\usepackage{textcomp}
\usepackage{multirow}
\usepackage{lipsum}
\usepackage{subcaption}
\usepackage{xcolor}
\usepackage{bm}
\usepackage{float}
\usepackage{comment}
\usepackage{todonotes}

\title{\LARGE \bf
Few-shot model-based adaptation in noisy conditions
}

\author{Karol Arndt$^{1}$, Ali Ghadirzadeh$^{1,2}$, Murtaza Hazara$^{1,3}$ and Ville Kyrki$^{1}$
\thanks{This work was supported by Academy of Finland grants 313966 and 317020. We also gratefully acknowledge the support of NVIDIA Corporation with the donation of the Titan Xp GPU used for this research. The authors also wish to acknowledge CSC – IT Center for Science, Finland, for computational resources.}
\thanks{$^{1}$Intelligent Robotics Group, Aalto University, Espoo, Finland
        {\tt\small first.last@aalto.fi}}%
\thanks{$^{2}$KTH Royal Institute of Technology, Stockholm, Sweden
        {\tt\small algh@kth.se}}%
\thanks{$^{3}$Department of Mechanical Engineering, KU Leuven, Belgium
        {\tt\small murtaza.hazara@kuleuven.be}}%
}

\begin{document}

\maketitle
\thispagestyle{empty}
\pagestyle{empty}

\begin{abstract}
Few-shot adaptation is a challenging problem in the context of simulation-to-real transfer in robotics, requiring safe and informative data collection.
In physical systems, additional challenge may be posed by domain noise, which is present in virtually all real-world applications.
In this paper, we propose to perform few-shot adaptation of dynamics models in noisy conditions using an uncertainty-aware Kalman filter-based neural network architecture.
We show that the proposed method, which explicitly addresses domain noise, improves few-shot adaptation error over a blackbox adaptation LSTM baseline, and over a model-free on-policy reinforcement learning approach, which tries to learn an adaptable and informative policy at the same time.
The proposed method also allows for system analysis by analyzing hidden states of the model during and after adaptation.
\end{abstract}

\section{Introduction}
Applying machine learning techniques to robot control tasks is a challenging problem, largely due to low sample efficiency of most machine learning methods.
In the case of reinforcement learning, additional challenge lies in the random exploration process, which takes place during learning, and poses a major risk of hardware damage.

A popular solution lies in using artificial data~\cite{hamalainen2019affordance,Tobin2017domain}
or physics simulators~\cite{andrychowicz19learning}
to facilitate the training; yet, in many cases, the simulation is not accurate enough for the model to achieve optimal performance in the real world without additional adaptation~\cite{transferMurtaza,arndt2019meta,Nagabandi2019learning}.

While few-shot learning has been quite extensively studied by the machine learning community, especially within the framework of meta-learning~\cite{schmidhuber:1987:srl,finn2017maml}, most prior works assume that noise-free labels are available for the learner during the adaptation process.
However, certain real-world physical systems can be noisy or stochastic;
that is, executing the same action in the same state may result in different observations.
In such scenarios, performing few-shot adaptation to real conditions may pose an additional challenge.
It has been shown that modelling and utilization of uncertainty information is effective in guiding and speeding up the policy learning for novel robotic tasks~\cite{8461241}; yet, in many real-world problems, the noise characteristics are not known in advance and thus cannot be easily injected into the simulated data.

\begin{figure}
    \centering
    \begin{subfigure}{.7\linewidth}
    \includegraphics[width=\textwidth]{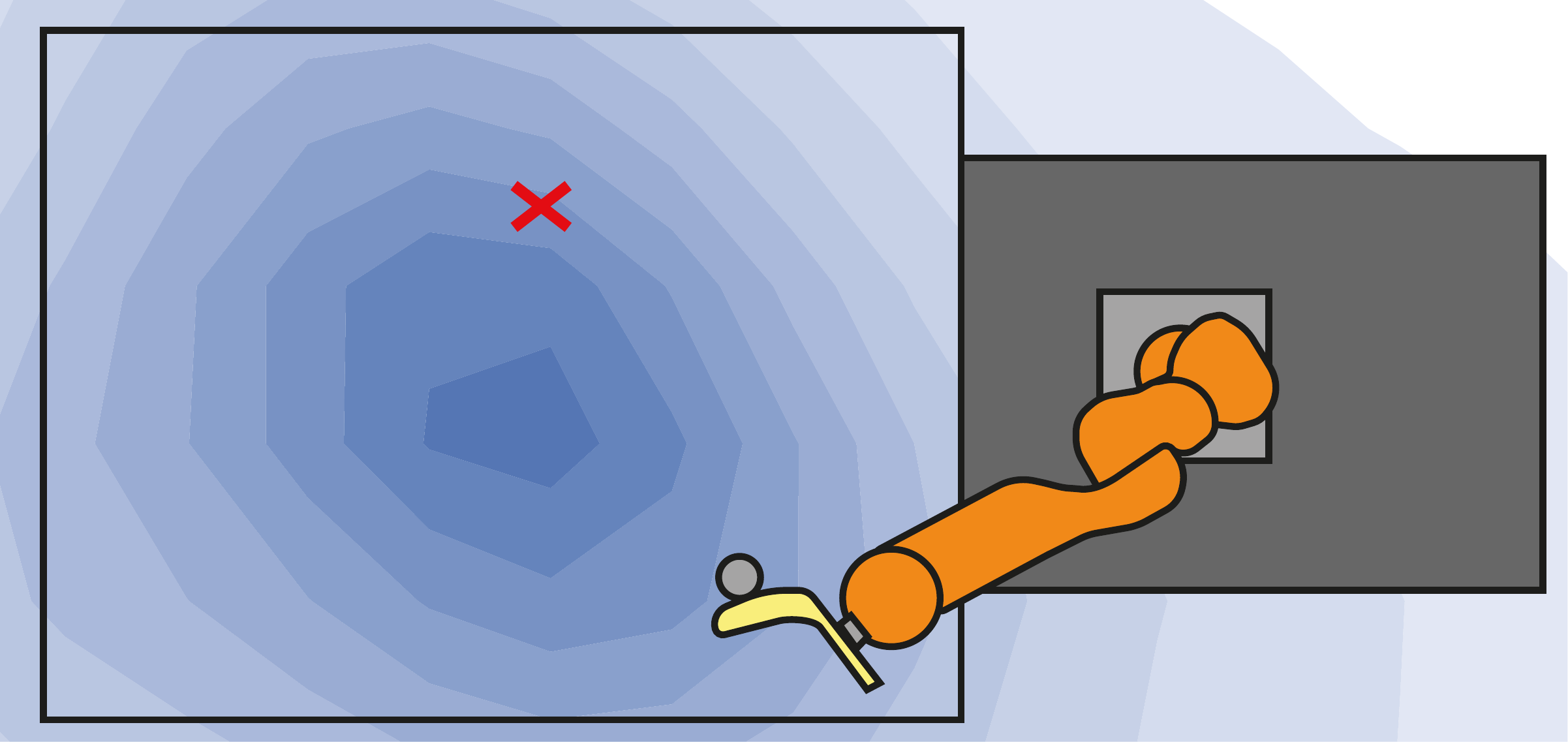}
    \caption{}
    \label{fig:intro_1}
    \end{subfigure}
    \begin{subfigure}{.7\linewidth}
    \includegraphics[width=\textwidth]{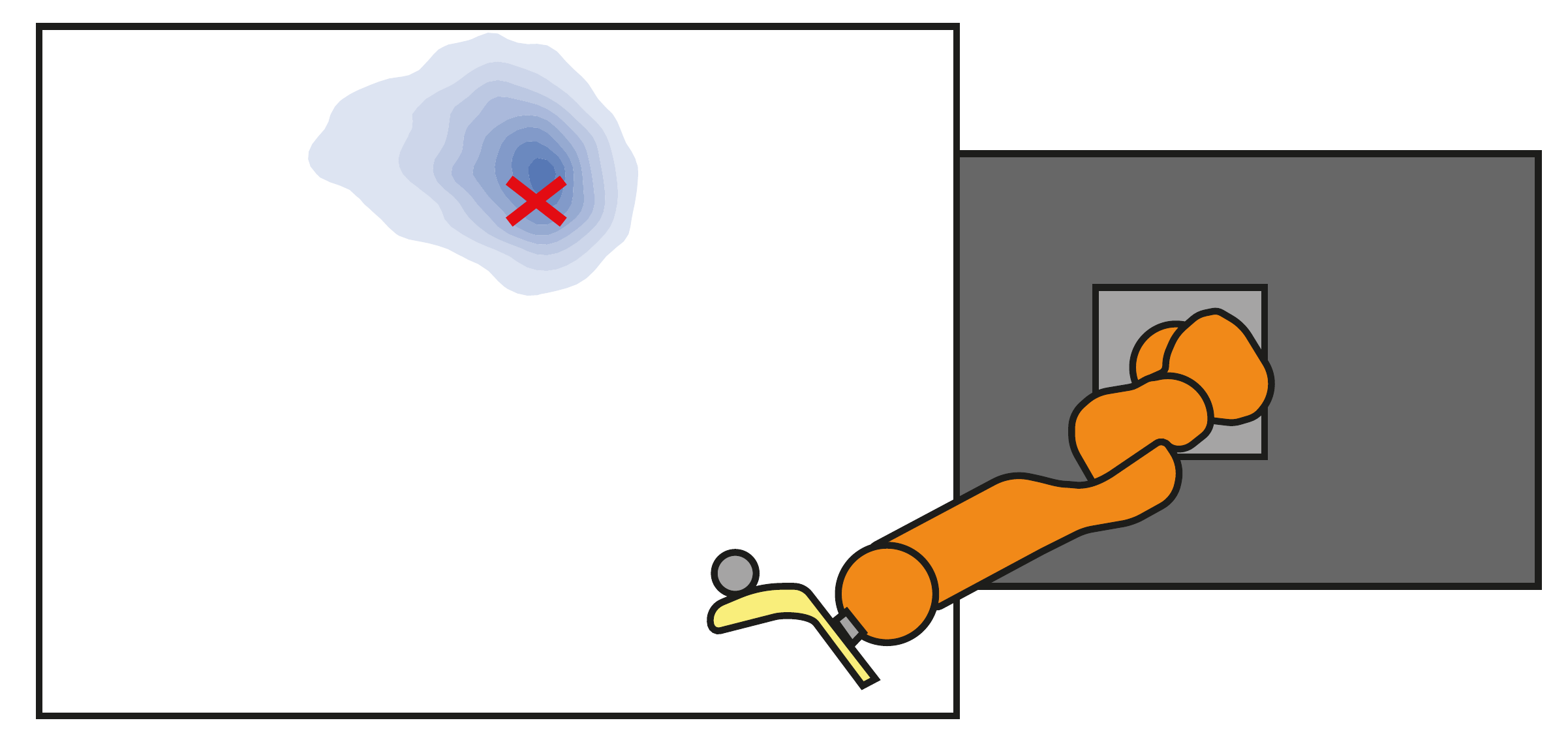}
    \caption{}
    \label{fig:intro_2}
    \end{subfigure}
    \caption{Before seeing any samples, the model produces large uncertainty about the final puck position~(\subref{fig:intro_1}). After performing some actions in the environment, the prediction uncertainty falls down~(\subref{fig:intro_2}). The red cross marks the true position.}
    \label{fig:intro}
    \vspace{-12pt}
\end{figure}

In this paper, we present an uncertainty-aware meta-learning approach to adapt a dynamics model trained across a variety of conditions in simulation to the physical world.
Our goal is to enhance sim-to-real transfer of dynamic skills in terms of data-efficiency and speed of adaptation through uncertainty-awareness.
In contrast to previously explored gradient-based methods~\cite{finn2017maml,arndt2019meta,Nagabandi2019learning}, which merely look at the mean gradient direction (ignoring the variance information), we propose to use a memory-based approach, which allow the model to keep track of uncertainty statistics~\cite{ortega2019metalearning}.
This uncertainty can be projected directly onto the predictions, as shown in Figure~\ref{fig:intro}; before adaptation (Figure~\ref{fig:intro_1}) the model reports large uncertainty about the position of the hockeypuck after a planned hit, while after adaptation the uncertainty falls down (Figure~\ref{fig:intro_2}).
In our method, this is achieved by using a network architecture designed around a trainable Kalman filter~\cite{kalman1960kalman,haarnoja2016backprop,becker2019recurrent}.

The focus on model adaptation, as opposed to policy adaptation, removes the need to run a random exploratory policy altogether;
instead, the data can be collected with the assistance of a human operator, or by a specifically designed policy which is known to be safe.
It also allows for the adapted environment model to be reused for different tasks in the same environment---namely, the same adapted state transition model can be used to maximize different objective functions in the environment, provided that the state-action space region relevant for the new task has been explored, or its properties have been identified based on prior information from the meta-training phase.

The contributions of this paper are as follows:
(1)~presenting a novel noise-aware meta-learning method based on a trainable Kalman filter,
(2)~showing that the proposed model structure outperforms LSTM and MAML on domain adaptation tasks in noisy conditions,
(3)~demonstrating that the learned latent representations of dynamic conditions are interpretable, corresponding to physical parameters,
(4)~showing that model adaptation through meta-learning is more data efficient compared to policy adaptation.

\section{Related work}

\subsection{Meta-learning}
Many meta-learning approaches are based on a recurrent neural network architecture, which the hidden state is updated based on the observed learning data~\cite{hochreiter2001learning}.
As the optimization procedure itself is learned together with the state representations, these methods are often referred to as \textit{blackbox meta-learning}.

With the advancement of deep learning, another family of meta-learning methods---optimization-based meta-learning---was introduced by Finn~et~al. with the model-agnostic meta-learning algorithm (MAML)~\cite{finn2017maml}.
Multiple extensions to the original algorithm were proposed later~\cite{flennerhag2019metalearning,stadie2018emaml}.
These methods, however, rely only on the mean gradient direction calculated over multiple samples while discarding the variance, and do not encode the uncertainty information in any way.
Various probabilistic extensions to MAML were introduced~\cite{finn2018platipus, yoon2018bmaml}; these approaches rely on ensemble models or Stein variational gradient descent to introduce sampling operations to the network during training, in order to encode uncertainty in the network output.


Our approach to model-based meta adaptation is based on the idea of training Kalman filters via backpropagation, as proposed in~\cite{haarnoja2016backprop}.
In that work, a Kalman filter is embedded inside a neural network and trained via backpropagation.
The architecture proposed in~\cite{haarnoja2016backprop} is shown to improve performance on standard state estimation tasks with deep learning over other recurrent architectures.
We propose to use a similar architecture for adaptation of dynamics models, rather than standard state estimation.

Recently, some considerations on uncertainty-aware meta-learning were put forward by Gordon~et~al.~\cite{gordon2018metalearning}, where it is proposed to view meta learning as learning a distribution over task-specific parameters.
Our method can be viewed as a special case of this generalized framework, where the parameters are modeled as Gaussian distributions and the inference is performed using standard Kalman filter update rules.

\subsection{Sim-to-real transfer in robotics}
The problem of sim-to-real transfer in robotics has been widely addressed in recent years, especially within the context of reinforcement learning. 
In the most basic approach, policies are trained in simulation and reused in the real world.
This, however, requires tedious fine-tuning of simulation parameters, and sometimes requires extreme measures, such as disassembling the robot and measuring the individual components in order to fine-tune the simulation~\cite{tan2018sim}.
Moreover, some physical phenomena such as backlash cannot be modeled in simulation, making the policies learned in simulation inefficient for real world.

The view-invariant servoing approach presented in~\cite{sadeghi17sim2real} can be thought of as an example of blackbox memory-based meta-learning; the recurrent model is trained over a variety of different simulated conditions such that it can adapt to real world conditions, as more and more samples are collected during operation.
This method, however, focuses on optimizing a task-specific policy, which requires that policy to be run on the physical setup before being exposed to any real world data.
Additionally, the collected data cannot be reused for other tasks, and the method does not account for the uncertainty present in real-world data.

Previous work on meta-learning for sim-to-real transfer focused on policy adaptation~\cite{Nagabandi2019learning,arndt2019meta,Song2020RapidlyAL}.
With these approaches, the final policy is a direct result of an on-policy update performed on the policy used to collect the data, as in~\cite{finn2017maml}.
As such, not only do the meta-policy parameters need to constitute a starting point for further adaptation, but also the meta-policy itself needs to explore the environment in a way which provides meaningful information for adaptation~\cite{stadie2018emaml}; there is, however, no straightforward and universal way of balancing between these two objectives~\cite{stadie2018emaml, gupta18metareinforcement, rothfuss2018promp}.
Using our method, on the other hand, a dynamic model can be optimized using data collected from a policy which is known to be safe, and is capable of providing informative samples.

Model adaptation with meta-learning has been previously utilized by Clavera~et~al.~\cite{clavera2018model} as a method of regularizing and stabilizing model-based reinforcement learning.
This method, however, only considered minor discrepancies between models, and---like other gradient-based methods---discarded uncertainty information in the update rule.
The method we propose in this paper, in contrast, aims at adapting to a wide variety of conditions and explicitly keeps track of the uncertainty (as expressed by the variance of the task-specific parameter vector).

\section{Method}
In this section, we first provide a formal statement of the problem we are addressing in this work.
We then proceed with a more detailed description of our approach to the presented problem.

\subsection{Problem formulation and solution overview}
Given a  state $s_i$ and an empty initial database $\mathcal{D}_0$ of state transitions $(s, a, s')$, we consider the problem of successively choosing one of $N$ actions. Each action $a_i$ leads to a new state $s_i'$. In addition, the state observation process may be disturbed by random noise, leading to noisy observation $\tilde{s}'$. Successively, we update the current database $\mathcal{D}_i = \mathcal{D}_{i-1} \cup \{(s_i, a_i, s_i')\}$ with the newly observed state action pair $(s_i, a_i, s_i')$. We assume that the dynamics of the considered sequential decision making problem can be modeled as a supervised learning problem:
\begin{equation}
    \begin{pmatrix}
    s_{i}' &&\hspace{-10pt} \phi_i
    \end{pmatrix}^\top
    = f_\theta(s_i, a_i, \phi_{i-1})
\end{equation}
using a function approximator $f_{\theta}(\cdot)$ where  $\theta$ denotes its parameters and $\phi_{i-1}$ represents the hidden state which is governed by a function $g$ of previous action and noisy state observations:
\begin{equation}
    \phi_{i-1} = g(\mathcal{D}_{i-1})
\end{equation}

In the proposed method, the hidden state $\phi$ is a task-specific parameter vector, describing the unknown system dynamics.

\begin{figure}
    \centering
    \includegraphics[width=\linewidth]{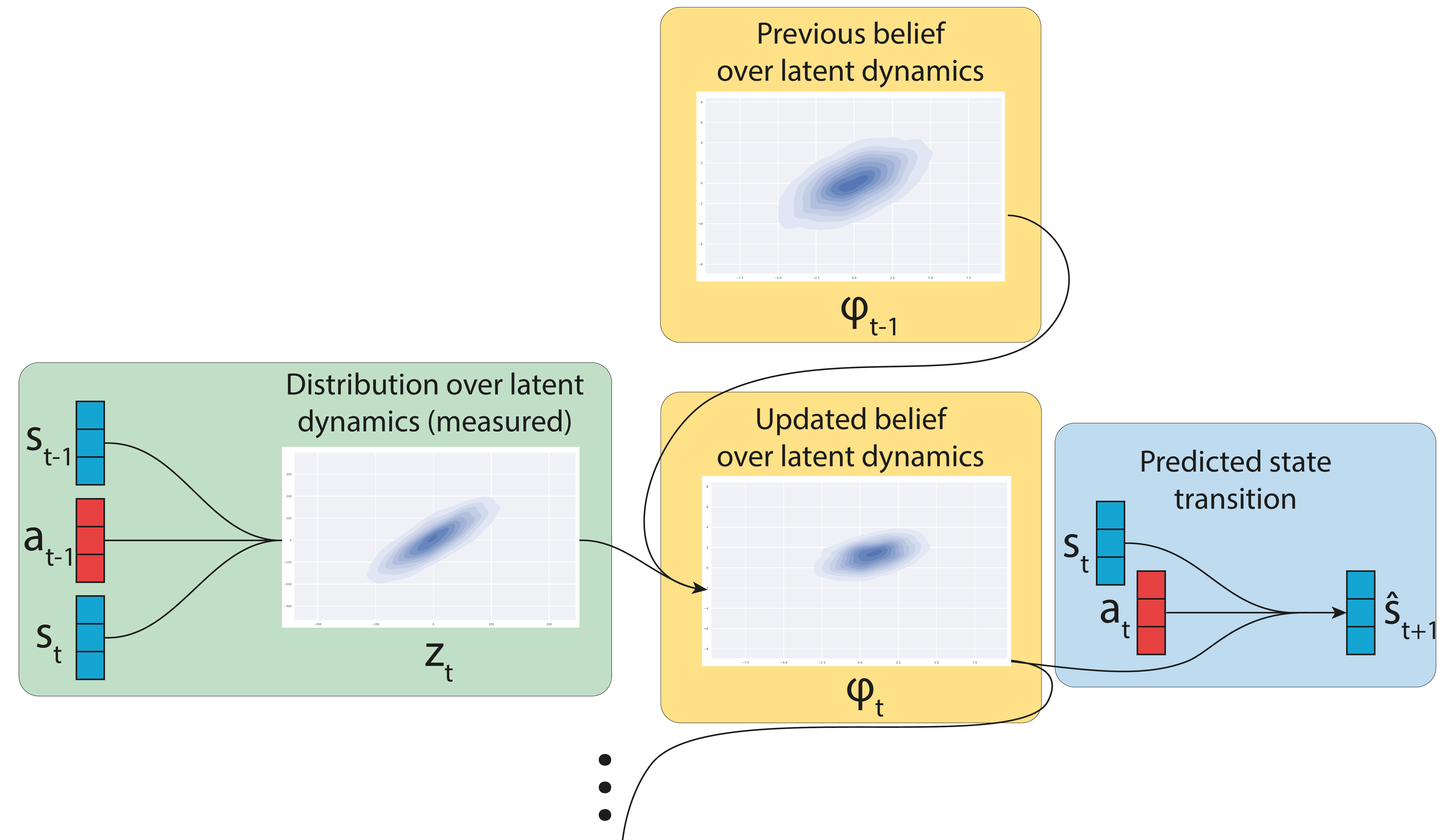}
    \caption{Method overview. The architecture consists of three main parts: measurement (green), state integration (yellow), and prediction (blue).}
    \label{fig:method_overview}
    \vspace{-12pt}
\end{figure}

In order to estimate this state from noisy measurements, we build a neural network, as shown in Figure~\ref{fig:method_overview}.
The architecture consists of three main parts: measurement (which estimates the system dynamics based on a noisy observation of a state transition), state integration (which integrates the parameter estimations together), and prediction (which estimates the next state given the previous state and the action).

\subsection{Measurement model}
The measurement model, shown in green in Figure~\ref{fig:method_overview}, estimates the distribution of possible system states $\tilde{z}$ (representing the dynamics of the system) given the measurement: $\tilde{z}_i \sim f^m_\theta(s_{i}, a_{i-1}, \tilde{s}'_{i-1})$. 
To make the system robust to observation noise, we corrupt the state observations with Gaussian noise: $\tilde{s} = s + \epsilon$, where $\epsilon \sim \mathcal{N}(0, \Sigma_s)$.
As we assume that we do not know the exact level of noise variance of the real environment, we incorporate $\Sigma_s$ into the task description, with $\Sigma_s$ being a diagonal matrix with values sampled from a uniform distribution:

\begin{equation} \label{eq:sigma_s}
    diag(\Sigma_s) \sim \mathcal{U}(0, \sigma^2_{s,max})
\end{equation}

The internal state representations are learned by the neural network in an unsupervised manner during training.
The distribution described by $f^m$ is modeled as a Gaussian parametrized by mean and covariance: $\tilde{z} = \mathcal{N}(\mu_z, \Sigma_z)$. 
This procedure corresponds to step~\ref{alg2:measure} of Algorithm~\ref{alg:metatraining}.

For many systems, it is impossible to uniquely determine the dynamic conditions based on a single measurement; i.e., the system can be seen as underdetermined from the parameter estimation perspective.
Thus, in addition to uncertainty introduced by measurement noise, the measurement covariance matrix has to convey the uncertainty caused by the possible underdetermined nature of the system.
We hence use heteroskedastic uncertainty, with covariance $\Sigma_z$ predicted for each measurement.
The returned measurement can thus be interpreted as a probability distribution over all systems from which the observed state transition could originate.

\subsection{Integration}
The measurement distribution returned by the measurement model is  passed to the recurrent part of the network, marked in yellow in Figure~\ref{fig:method_overview}, which is responsible for integrating the observations together.
In our method, this is achieved by a deep Kalman filter~\cite{haarnoja2016backprop}---a Kalman filter embedded within a neural network, where the parameter matrices of the filter are trained with backpropagation together with the rest of the network.
Such a network has the inference procedure built-in in the computation graph, which was shown to improve its ability to integrate information from individual samples and reason about uncertainty~\cite{haarnoja2016backprop}.

\SetKwInput{KwData}{Input}
\vspace{-4pt}
\begin{algorithm}
\SetAlgoLined
\KwData{Dataset $\{(s, a, s')\}$ of state transitions}
\KwResult{Network parameters $\theta$}
 randomly initialize network parameters $\theta$\;
 \While{not converged}{
    Reset internal parameters ($\mu_\phi$ and $\Sigma_\phi$)\; 
    Sample task $\tau$ (dynamics and noise variance,~Eq.~(\ref{eq:sigma_s}))\; \label{alg2:sample_task}
    Sample a sequence of $N$ state transitions $(s, a, s')$ from $\tau$\;  \label{alg2:sample_trans}
    \For{each transition $(s, a, s')$ in the sequence} {
        Generate noise: $\tilde{s'} \sim \mathcal{N}(s', \Sigma_s)$\;  \label{alg2:get_noise}
        Update state estimate (Eqs. \ref{eq:pred_mean} and \ref{eq:pred_cov})\;  \label{alg2:preupdate}
        Estimate the dynamic conditions $\mu_z, \Sigma_z = f^m_\theta(s, a, \tilde{s}')$ \;  \label{alg2:measure}
        Update $\mu_\phi$ and $\Sigma_\phi$ conditioned on $\mu_z$ and $\Sigma_z$ (Eqs. \ref{eq:kalmangain}, \ref{eq:mu_update}, \ref{eq:sigma_update})\;  \label{alg2:update}
        Predict $\hat{s'} = f_{\theta}^p(s, a, \mu_\phi)$\;  \label{alg2:predict}
    }
    Calculate loss over the sequence (Eq. \ref{eq:loss}) \; \label{alg2:loss}
    Calculate $\nabla_\theta \mathcal{L}$ and update $\theta$ using Adam \; \label{alg2:metaaupdate}
    \vspace{-4pt}
 }
 \caption{Training the dynamics model}
 \label{alg:metatraining}
\end{algorithm}

The first step of a Kalman filter performs a state prediction based on a learned model 
\begin{equation} \label{eq:pred_mean}
    \mu_{\phi}^{t+1|t} = A \mu_{\phi}^{t|t} + Bu,
\end{equation}
where $\mu_{\phi}^{t|t}$ represents the mean of the current belief about the state of the system, and $u$ represents external input.
In our case, where the state actually represents a dynamic parameter vector, the action $u$ would correspond to an external action which directly impacts the dynamics.
We assume that no such action takes place, and thus we set $B$ to zero.

The state estimation is modeled as a Gaussian with the covariance
\begin{equation} \label{eq:pred_cov}
    \Sigma_{\phi}^{t+1|t} = A^\top \Sigma_{\phi}^{t|t} A + Q .
\end{equation}

In the state dynamics prediction formulation, this update represents temporal change in dynamics.
For a stationary system, $A$ is the identity matrix, and a small non-zero value of the covariance matrix $Q$ can be used to approximate unmodeled drift in dynamic conditions and prevent the estimation covariance from approaching zero in limit, enabling lifelong learning.
This procedure corresponds to step~\ref{alg2:preupdate} of Algorithm~\ref{alg:metatraining}.

The state prediction is updated in step~\ref{alg2:update} of Algorithm~\ref{alg:metatraining} by integrating information coming from a measurement, following the standard Kalman filter equations~\cite{haarnoja2016backprop}, and resulting in $\mu_{\phi}^{t+1|t+1}$ and $\Sigma_{\phi}^{t+1|t+1}$.
First, the Kalman gain is calculated with

\begin{equation} \label{eq:kalmangain}
    K_t = \Sigma_{\phi}^{t+1|t} C_z^\top (C_z \Sigma_{\phi}^{t+1|t} C^\top_z + \Sigma_z)^{-1}
\end{equation}

Then, the belief about the mean $\mu_\phi$ is updated with

\begin{equation} \label{eq:mu_update}
    \mu_\phi^{t+1|t+1} = \mu_\phi^{t+1|t} + K_t (z_t - C_z \mu_\phi^{t+1|t})
\end{equation}

Finally, the covariance of the state belief is updated by

\begin{equation} \label{eq:sigma_update}
    \Sigma_{\phi}^{t+1|t+1} = (I - K_t C_z) \Sigma_{\phi}^{t+1|t}
\end{equation}

We use a standard linear Kalman filter, as the non-linear correspondence between the measurements and internal states is already addressed by the non-linear measurement system.
The values of $C_z$ and the initial state distribution (parametrized by $\mu_\phi^0$ and $\Sigma_\phi^0$) are learned during outer-loop optimization.
The use of Kalman filters allows us to introduce additional information in the prediction model---namely, we assume that the system is stationary (the transition matrix $A=I$), while allowing for some temporal drift to encourage the system to keep adapting over long periods of time and to improve numerical stability ($Q=\epsilon I$).

\subsection{State prediction}
Finally, the future state of the environment is predicted for the queried state-action pair by the \textit{prediction model} $\hat{s}'_i = f^p_\theta(s_i, a_i, \mu_\phi)$.
This model, like the measurement model, is a neural network.
The state prediction takes place in step~\ref{alg2:preupdate} of Algorithm~\ref{alg:metatraining}.

During meta-training on data collected in a simulated environment, the optimization objective $\mathcal{L}$ is calculated as the mean-squared error between the predictions of the entire model at each timestep over the whole sequence, and the noiseless ground-truth observations $s_i'$:
\begin{equation} \label{eq:loss}
    \mathcal{L} = \frac{1}{N} \sum_{i=1}^{N} (\hat{s_i}' - s_i')^2.
\end{equation}
The loss calculation and model optimization are performed, respectively, in steps~\ref{alg2:loss} and~\ref{alg2:metaaupdate} of Algorithm~\ref{alg:metatraining}.

The model is trained as a whole in an end-to-end fashion.

\section{Experiments}
In this section, we provide the details of the experimental evaluation of our method and present the results.
In order to illustrate the details of the adaptation process, we first study the performance of the method on a simple regression problem with a step-by-step walkthrough.
For the sake of demonstration, we visualize how the state and output distributions are changing as more samples are observed. 
Then, we evaluate the method's ability to adapt to different simulated conditions with different noise levels in MuJoCo on FetchSlide, an environment from OpenAI Gym~\cite{openaigym}.
Finally, we evaluate the sim-to-real performance of the method on a hockeypuck hitting task, using simulated data for training and data collected from a real robot for adaptation and evaluation.

The comparison is performed both against gradient-based adaptation methods~\cite{finn2017maml,arndt2019meta} and against an LSTM baseline~\cite{hochreiter2001learning}, which uses a blackbox adaptation scheme and does not explicitly account for noise.

\subsection{Linear regression}
In order to illustrate how the internal state $\phi$ and the prediction are changing during the adaptation process on a basic example, we devise a simple regression problem, where the goal is to predict values of a linear function based on a small amount of noisy observations $(x_i, \tilde{y}_i)$.
In this setup, instead of taking state-action pairs $(s, a)$ and predicting future states $\hat{s}'$, the model receives $x$ as input and predicts the value of the function, $\hat{y}$.

\begin{figure}
    \centering
    \begin{subfigure}{0.45\linewidth}
    \includegraphics[width=\linewidth]{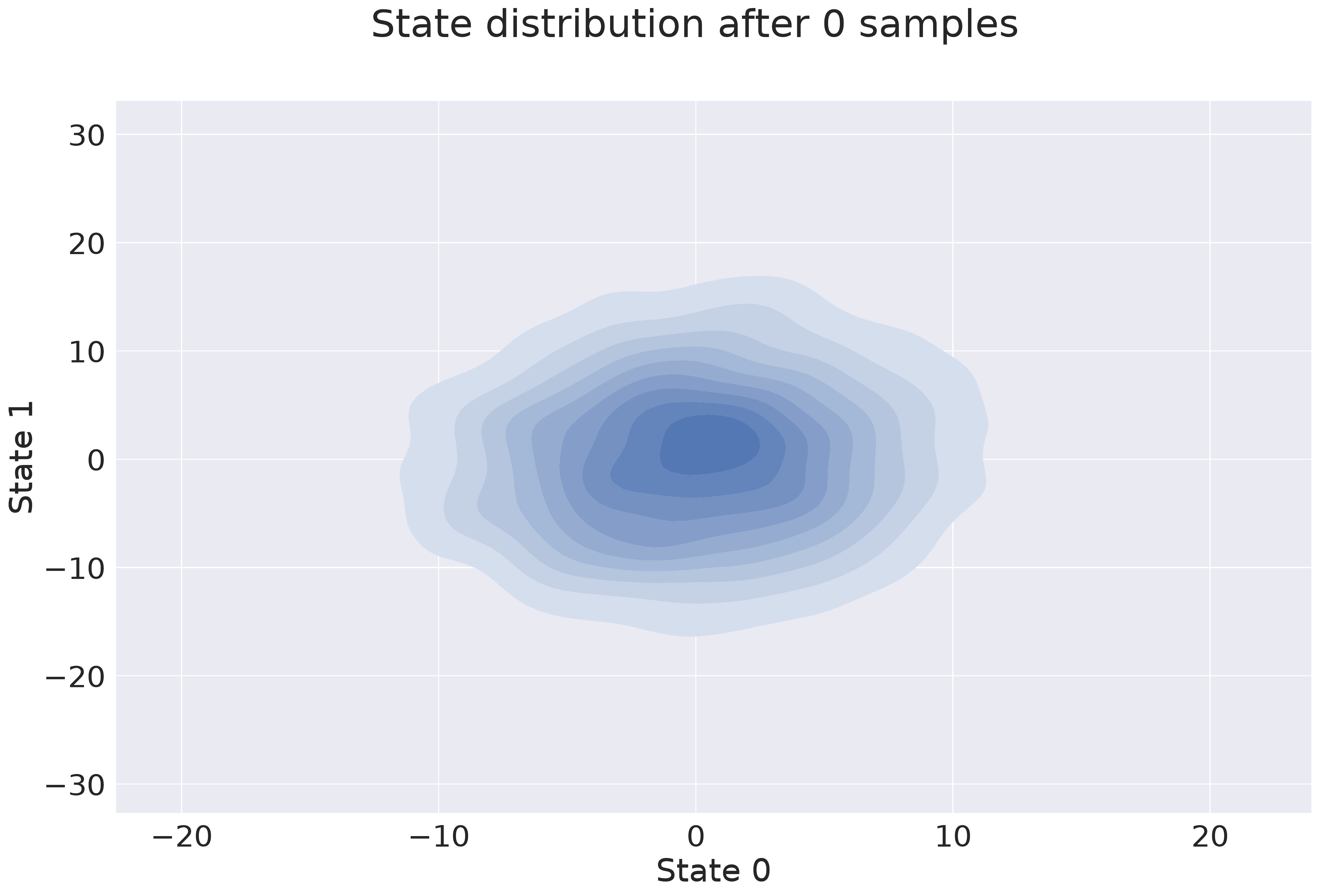}
    \caption{}
    \label{fig:linreg_d_s0}
    \vspace{-2pt}
    \end{subfigure}
    \begin{subfigure}{0.45\linewidth}
    \includegraphics[width=\linewidth]{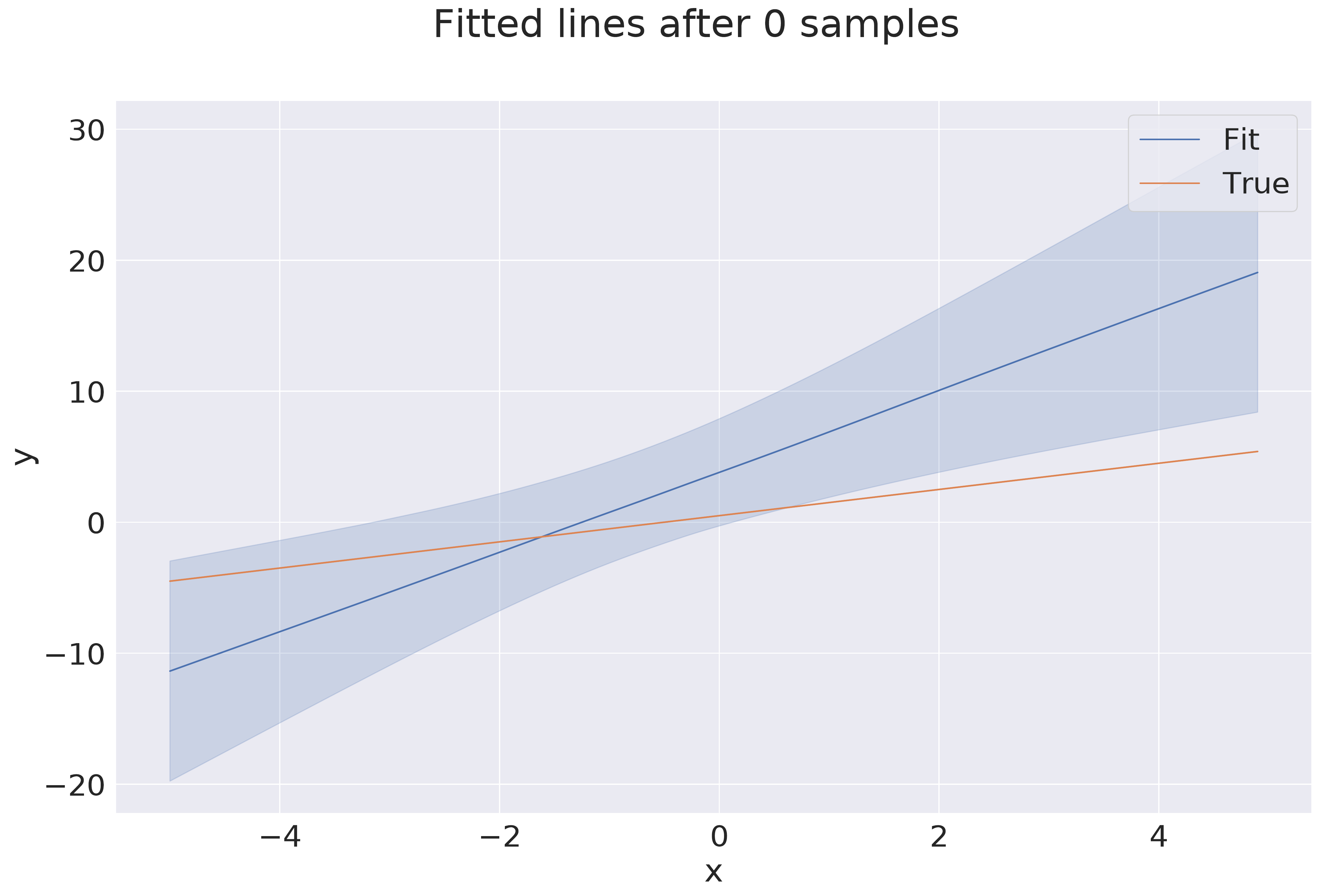}
    \caption{}
    \label{fig:linreg_d_y0}
    \vspace{-2pt}
    \end{subfigure}
    
    \begin{subfigure}{0.45\linewidth}
    \includegraphics[width=\linewidth]{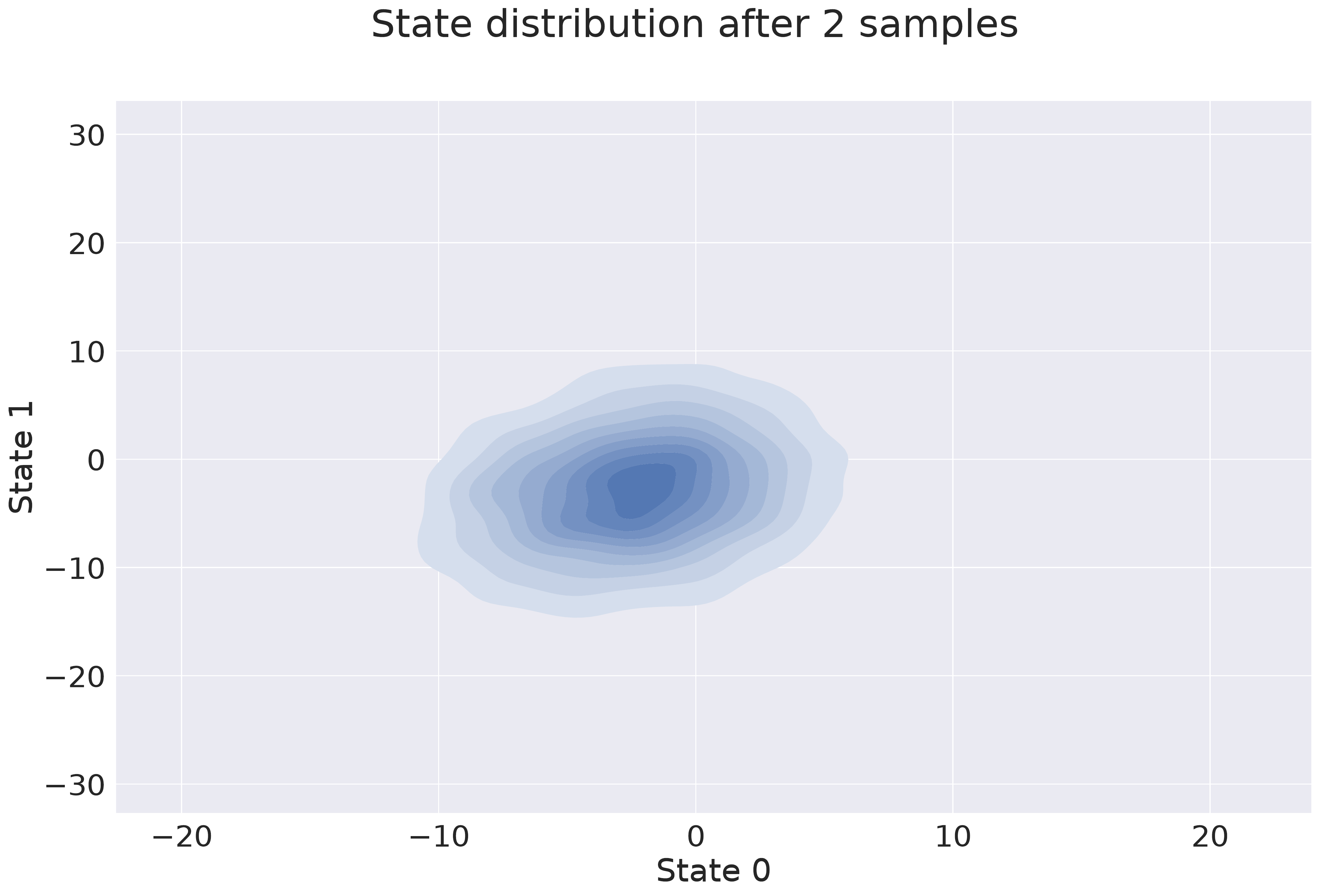}
    \caption{}
    \label{fig:linreg_d_s2}
    \vspace{-2pt}
    \end{subfigure}
    \begin{subfigure}{0.45\linewidth}
    \includegraphics[width=\linewidth]{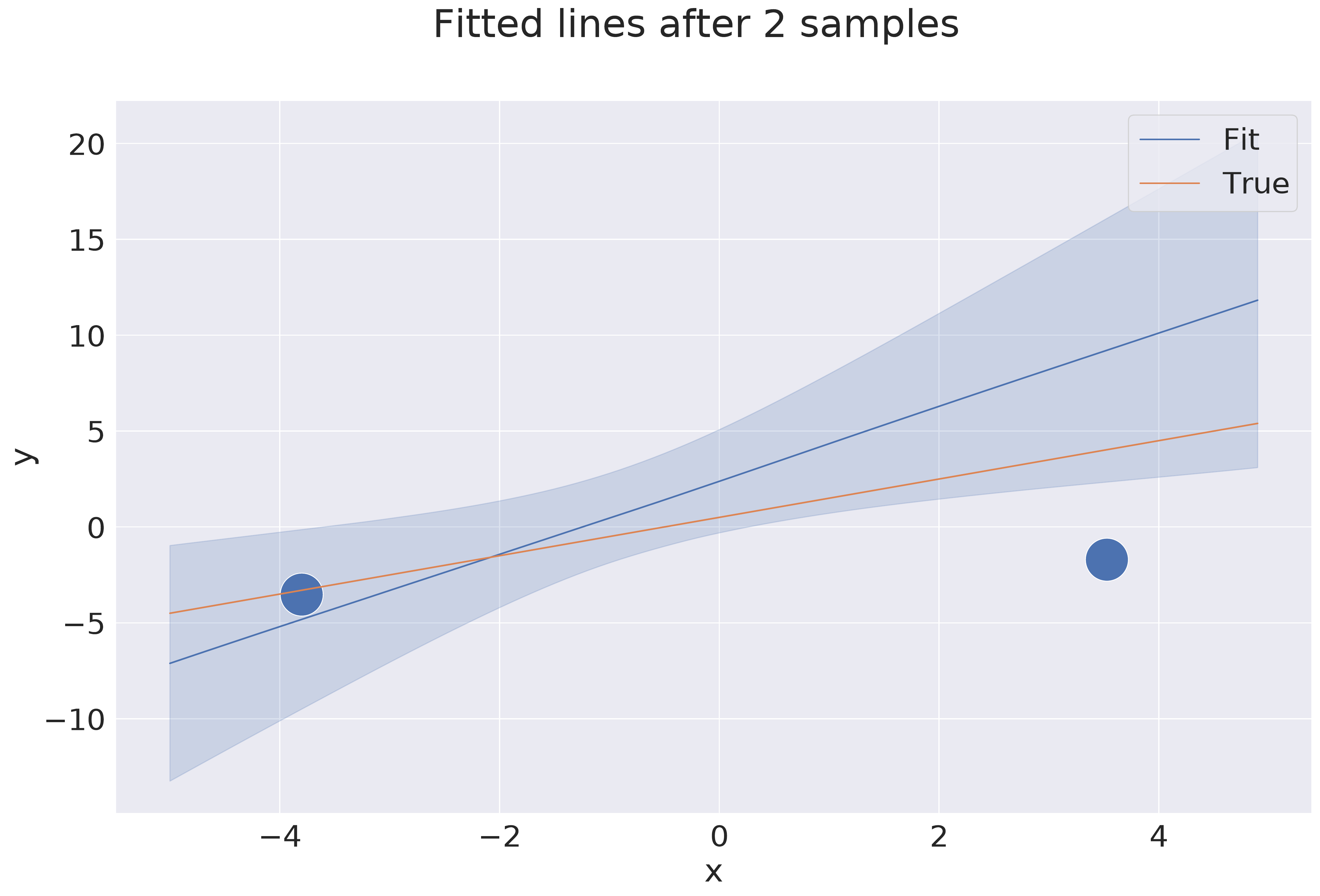}
    \caption{}
    \label{fig:linreg_d_y2}
    \vspace{-2pt}
    \end{subfigure}
    
    \begin{subfigure}{0.45\linewidth}
    \includegraphics[width=\linewidth]{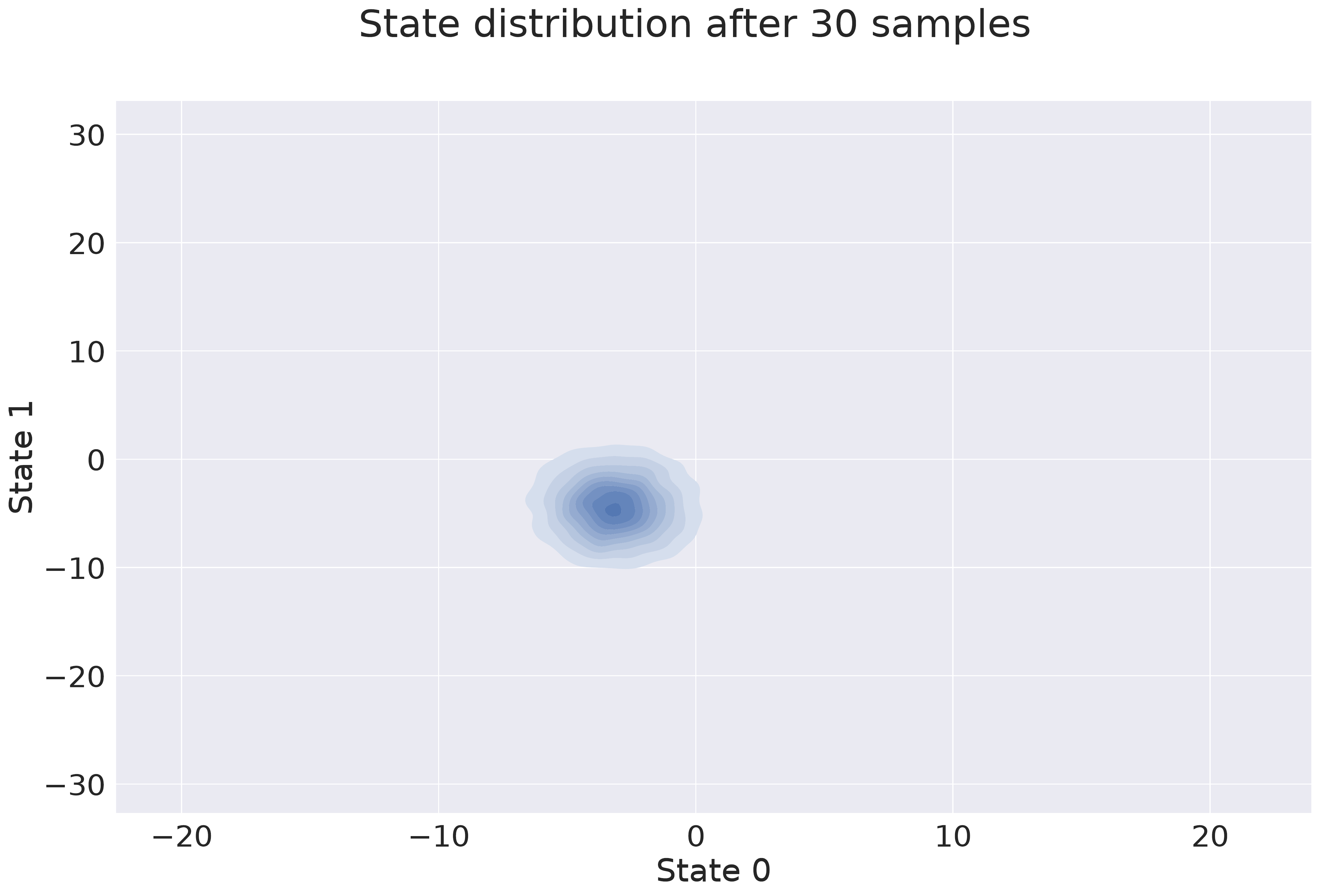}
    \caption{}
    \label{fig:linreg_d_s10}
    \vspace{-2pt}
    \end{subfigure}
    \begin{subfigure}{0.45\linewidth}
    \includegraphics[width=\linewidth]{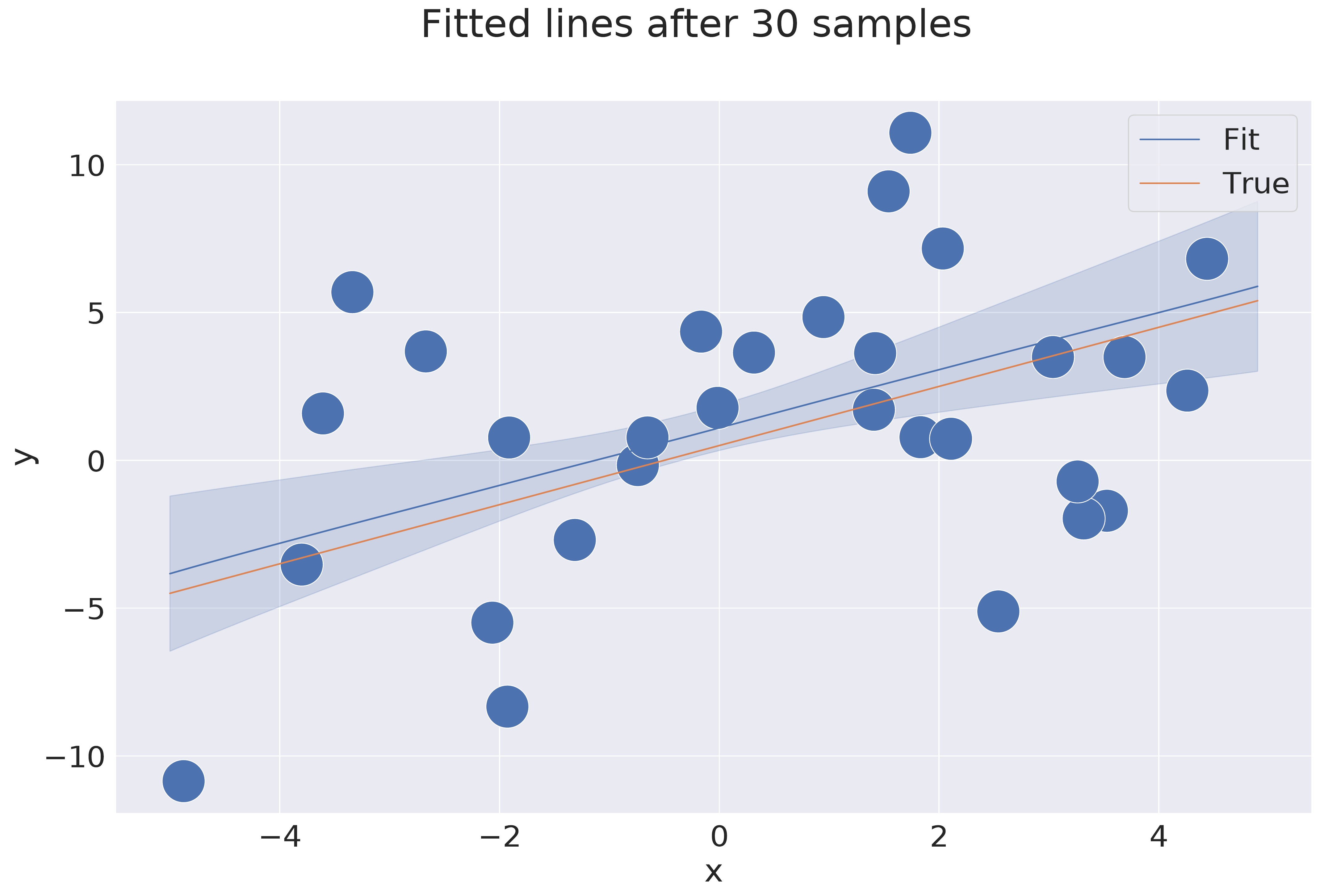}
    \caption{}
    \label{fig:linreg_d_y10}
    \vspace{-2pt}
    \end{subfigure}
    \vspace{-2pt}
    \caption{Analysis of adaptation mechanism, visualized on linear regression. The blue dots in (\subref{fig:linreg_d_y2}) and (\subref{fig:linreg_d_y10}) show the observed samples.}
    \label{fig:linreg_indepth}
    \vspace{-8pt}
\end{figure}

The changes in state and prediction distributions are shown in Figure~\ref{fig:linreg_indepth}.
Figure~\ref{fig:linreg_d_s0} shows the prior state distribution over parameters.
This distribution is learned during outer-loop optimization in the meta-learning phase and represents the prior over all dynamics conditions seen during training, which minimizes the expected prediction error over all samples, before any data is observed.
This behaviour arises as a result of including the initial prediction error in the optimization objective (Eq.~\ref{eq:loss}).
We sample values of the output $\hat{y}$ by sampling hidden state values $\phi$ from $\mathcal{_\phi, \Sigma_\phi}$ and passing the sampled hidden state values to the prediciton model.
This has been shown in Figure~\ref{fig:linreg_d_y0}).
The shaded area shows one standard deviation in state space projected onto the output domain.

After two samples are obtained, the variance of the latent variable noticeably goes down (Figure~\ref{fig:linreg_d_s2}).
Due to noise, the observed points do not exactly align with the true function, and the predicted line does not exactly go through the observed points; rather, it is still influenced by the prior (Figure~\ref{fig:linreg_d_y2}), as represented as the initial state belief.

As more samples are observed, the state and prediction variance further goes down and the prediction is more accurate (Figures~\ref{fig:linreg_d_s10} and \ref{fig:linreg_d_y10}), despite the observations being very noisy.
The impact of the prior is also reduced as more samples are observed.
Due to non-zero value of $Q$, the prediction uncertainty will never go down to zero, allowing for lifelong adaptation to changing conditions.

\subsection{FetchSlide}
\begin{figure}
    \centering
    \begin{subfigure}{0.49\linewidth}
    \includegraphics[width=\linewidth]{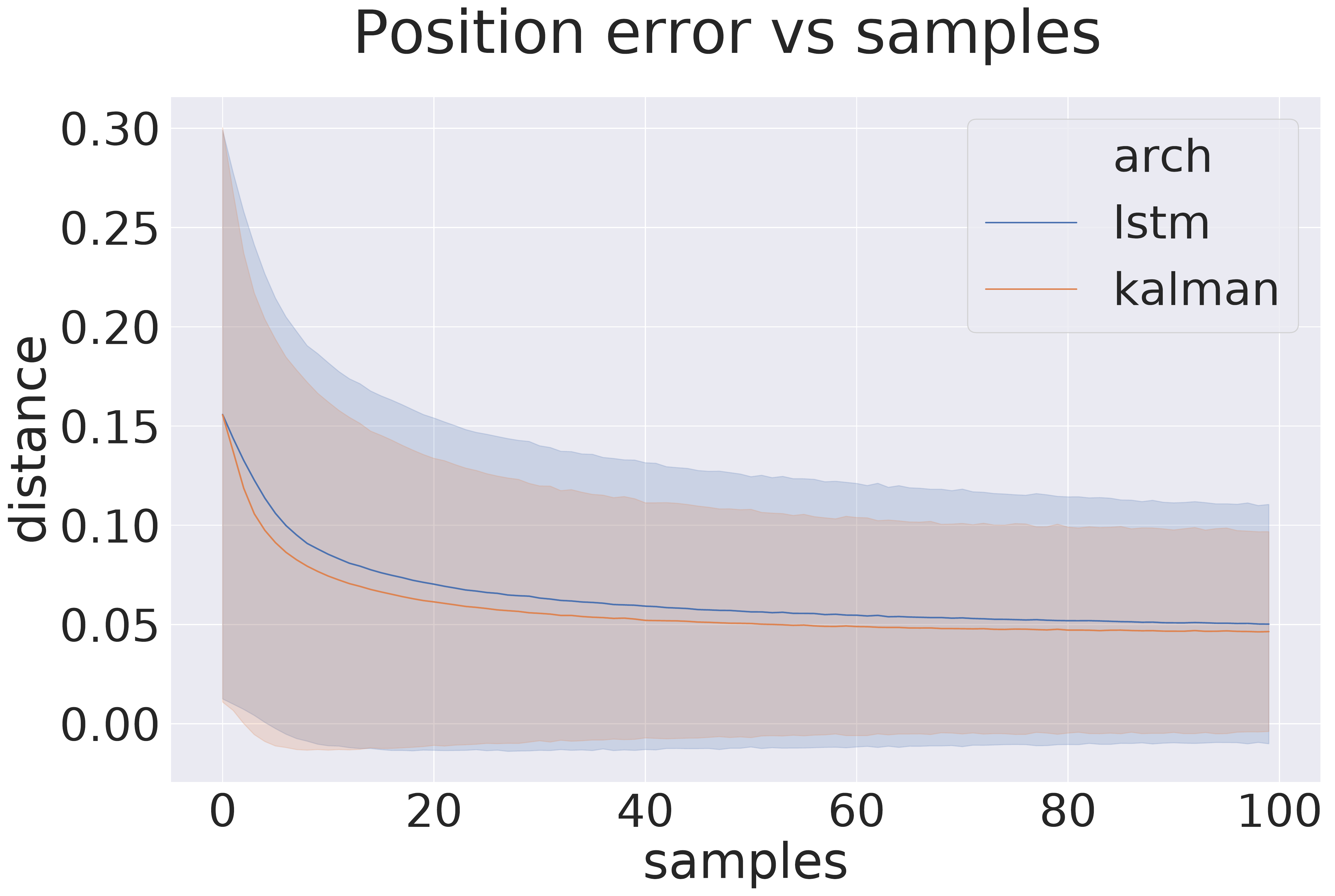}
    \vspace{-2pt}
    \end{subfigure}
    \begin{subfigure}{0.49\linewidth}
    \includegraphics[width=\linewidth]{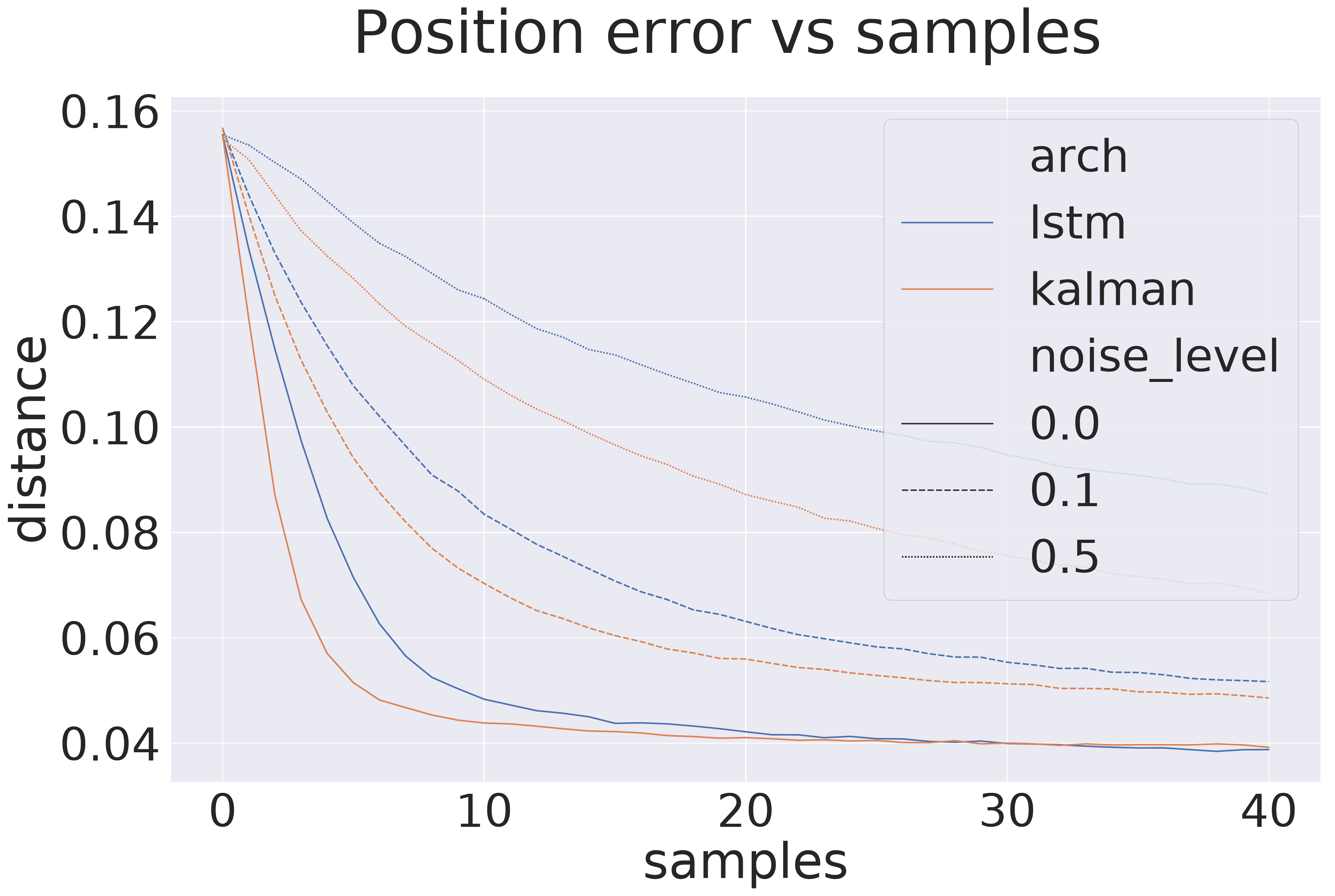}
    \vspace{-2pt}
    \end{subfigure}
    \caption{Domain adaptation results in FetchSlide}
    \label{fig:fs_res}
    \vspace{-8pt}
\end{figure}

\begin{figure}
    \centering
    \begin{subfigure}{0.49\linewidth}
    \includegraphics[width=\linewidth]{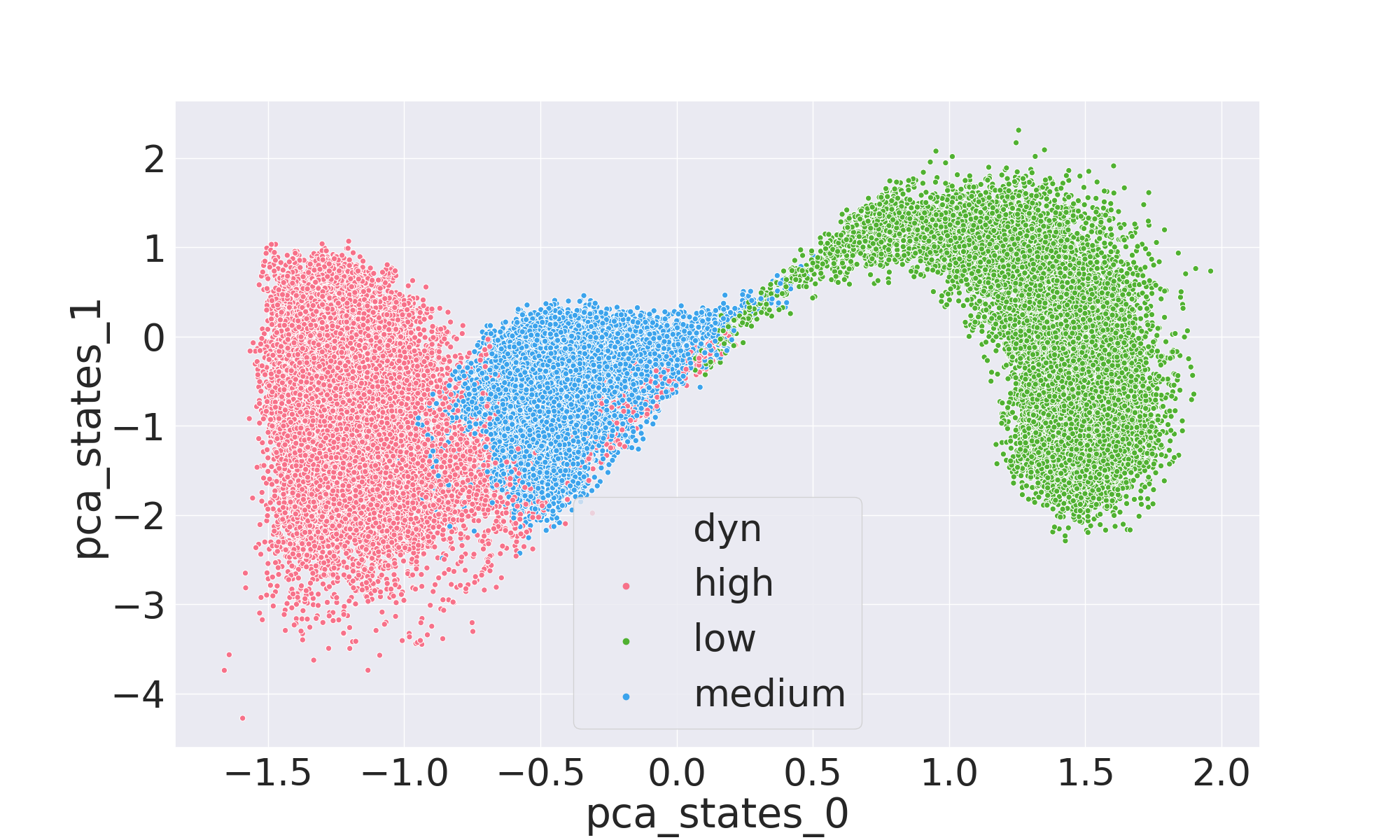}
    \caption{}
    \label{fig:fs_pca_kal_dyn}
    \vspace{-2pt}
    \end{subfigure}
    \begin{subfigure}{0.49\linewidth}
    \includegraphics[width=\linewidth]{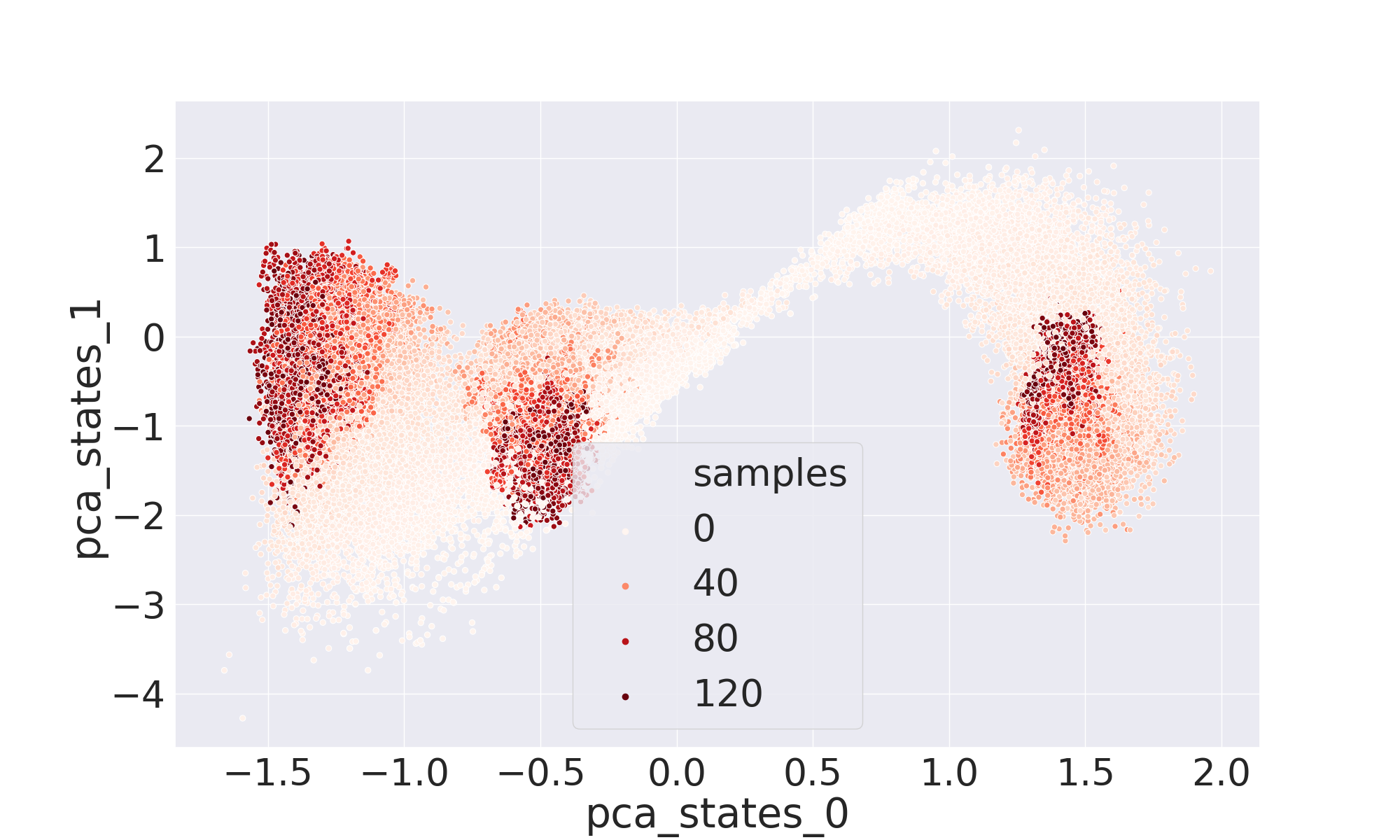}
    \caption{}
    \label{fig:fs_pca_kal_samples}
    \vspace{-2pt}
    \end{subfigure}
    \begin{subfigure}{0.49\linewidth}
    \includegraphics[width=\linewidth]{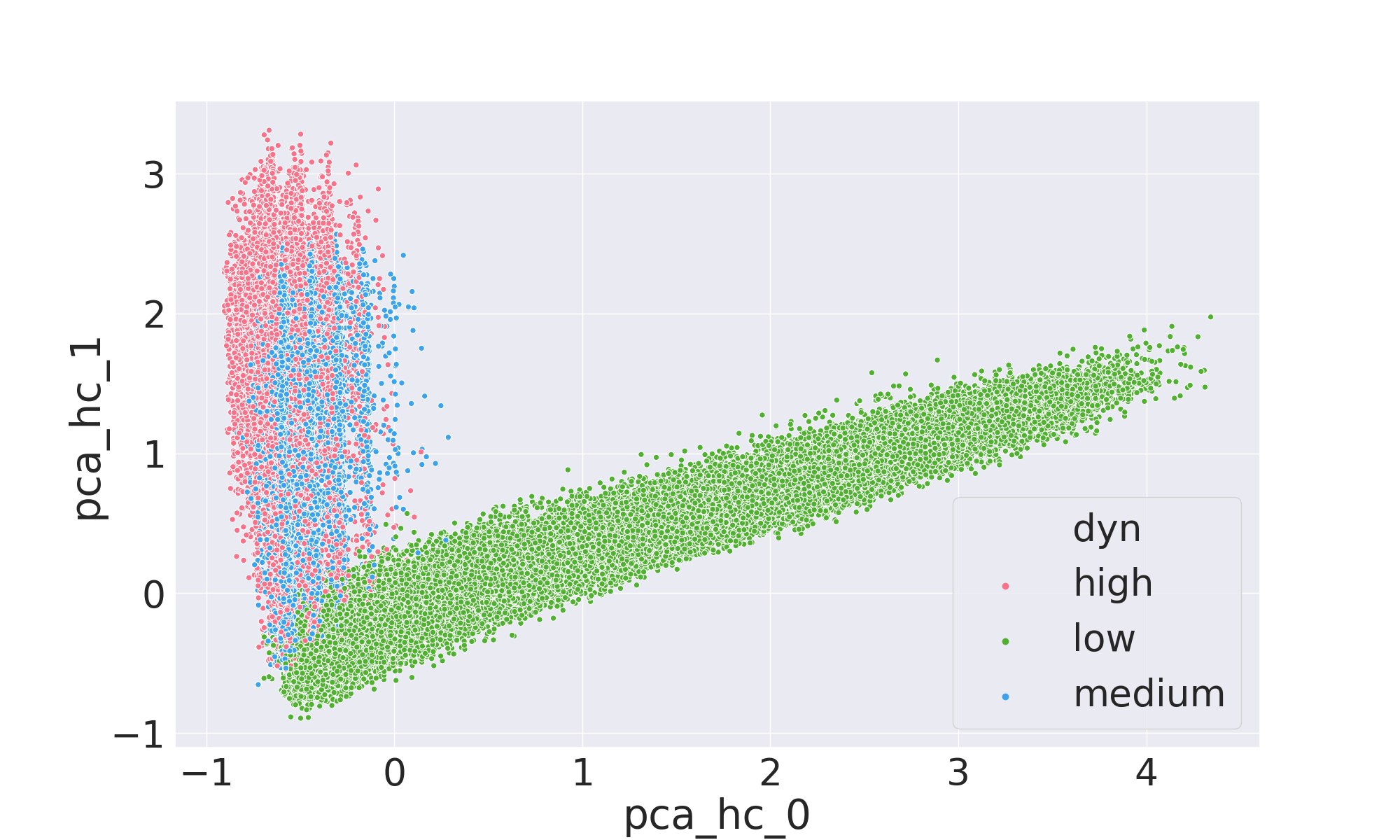}
    \caption{}
    \label{fig:fs_pca_lstm_dyn}
    \vspace{-2pt}
    \end{subfigure}
    \begin{subfigure}{0.49\linewidth}
    \includegraphics[width=\linewidth]{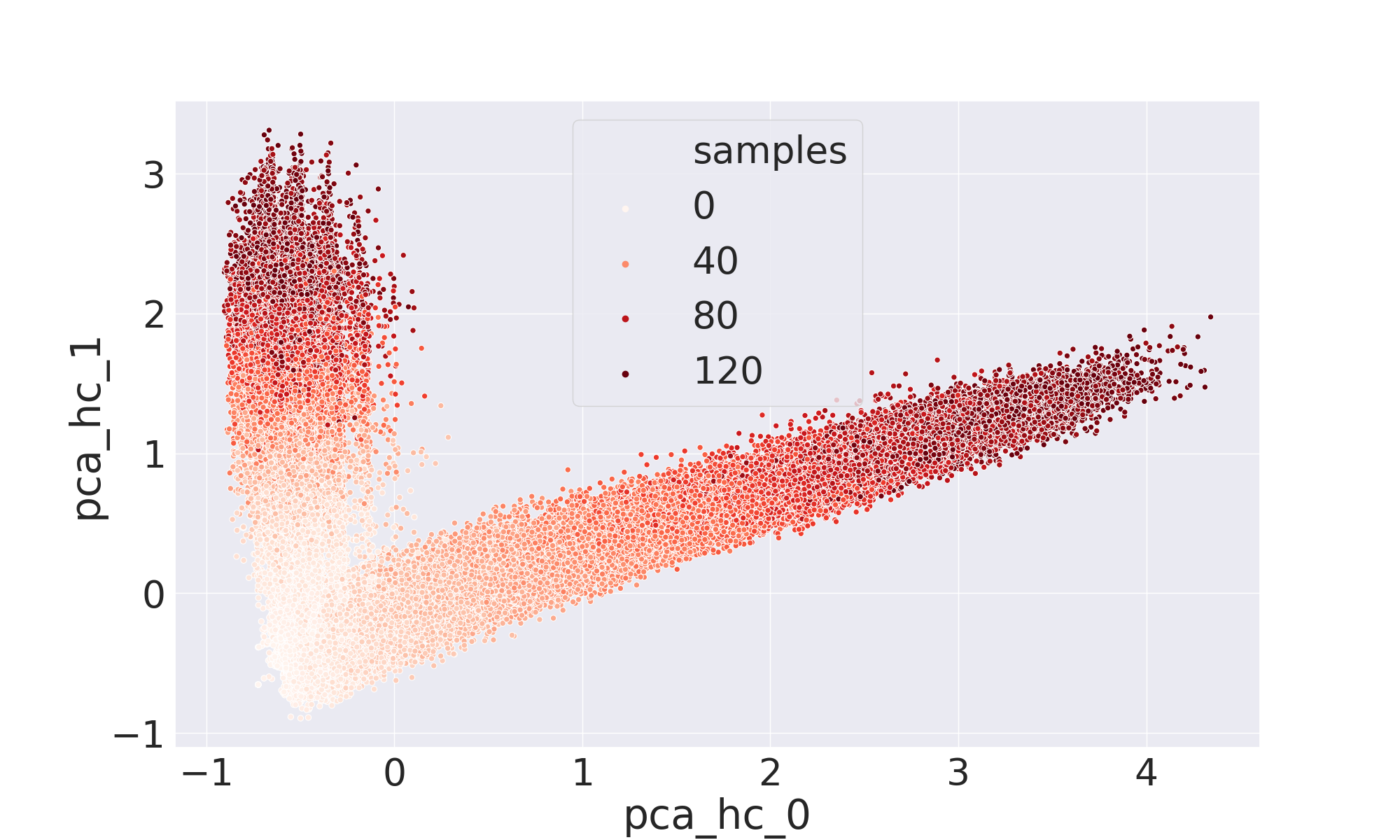}
    \caption{}
    \label{fig:fs_pca_lstm_samples}
    \vspace{-2pt}
    \end{subfigure}
    \caption{PCA analysis of hidden states in FetchSlide for the proposed (\subref{fig:fs_pca_kal_dyn}, \subref{fig:fs_pca_kal_samples}) and LSTM-based (\subref{fig:fs_pca_lstm_dyn}, \subref{fig:fs_pca_lstm_samples}) model adaptation}
    \label{fig:fs_pca_res}
    \vspace{-8pt}
\end{figure}

\begin{table}[]
    \centering
    \begin{tabular}{c|c}
        Parameter & Distribution \\
        \hline
        Linear friction & $\mathcal{U}(0.05, 0.7)$ \\
        Torsional friction & $\mathcal{U}(10^{-4}, 10^{-2})$ \\
        Rotational friction & $\mathcal{U}(5\cdot 10^{-5}, 10^{-3})$ \\
    \end{tabular}
    \vspace{-2pt}
    \caption{FetchSlide randomization parameters}
    \label{tab:fs_params}
    \vspace{-8pt}
\end{table}

To verify that the method is suitable for domain adaptation in noisy conditions, we built an experimental setup based on the FetchSlide environment from OpenAI Gym~\cite{openaigym}.
In this environment, the goal is to hit an object located at a random reachable location on the table such that it slides to a given target location.
In addition to the robot joint position, both the start and target locations are known to the agent and included in the state vector.
The robot is controlled via velocity commands given in Cartesian space.
We used 16 dimensions for the measurement and dynamics representations $z$ and $\phi$.

Following prior work on using generative models and latent space trajectory representations~\cite{ghadirzadeh2017deep,hamalainen2019affordance,arndt2019meta}, we use a variational autoencoder trained on a set of task-specific trajectories provided by human experts.
This approach removes the time complexity from the problem and provides a straightforward way of collecting data for model adaptation---as the latent distribution is known to be a unit Gaussian, new trajectories can be generated by simply sampling latent vectors from that distribution and passing the latent values to the decoder.
We trained the trajectory model on a set of 5.6 million trajectories which move the arm to a random point on the table and push the object away from the robot with a random angle and velocity.

Using the trajectory model, we collected a set of 6.93 million hits under 28875 random friction conditions in simulation.
The parameters used for friction distribution are presented in Table~\ref{tab:fs_params}.
For both the baseline and the proposed method, we trained the final dynamics model on sequences of 100 samples.
We used $\sigma_{s,max}^2= 0.3$.
For each method, we trained four separate models with Adam~\cite{Kingma14Adam} and averaged the test results.

The evaluation data was collected in the simulator under three scenarios---low, medium, and high friction. 
During evaluation, we additionally consider three scenarios---no observation noise, low noise ($\sigma^2=0.1$) and high noise ($\sigma^2=0.5$), where the high noise condition corresponds to larger noise variance than was seen in training.
The results of this evaluation are shown in Figure~\ref{fig:fs_res}.

We see that the proposed method outperforms the baseline most notably in the low-sample regime and high noise conditions.
Improved ability to generalize to out-of-distribution tasks can be attributed to encoding optimization within the computation graph, similarly to how MAML improves generalization performance through embedding gradient descent~\cite{finn2018universality}.

We additionally analysed the principal components (PCs) of the hidden states $\phi$ in both networks.
The results are shown in Figure~\ref{fig:fs_pca_res}.
The left subplots (\ref{fig:fs_pca_kal_dyn} and \ref{fig:fs_pca_lstm_dyn}) are color-coded by the evaluation domain (low, medium or high friction) and the right ones (\ref{fig:fs_pca_kal_samples} and \ref{fig:fs_pca_lstm_samples}) by the number of samples observed by the model.
Looking at Figure~\ref{fig:fs_pca_kal_dyn}, we can observe that the dynamics lie in the order of decreasing friction, with high friction on the left, medium friction in the middle, and low friction much farther to the right. 

We also observe that the LSTM model encodes uncertainty, as expressed by the number of observed samples, in the hidden states.
Additionally, it does not clearly separate the medium and high friction conditions; rather, the PCA projections span the same area even after observing 80 samples.
The proposed Kalman filter-based architecture clearly splits the two domains already with less than 20 samples.

\subsection{Hockey puck}
\begin{figure}
    \centering
    \begin{subfigure}{0.23\linewidth}
    \includegraphics[width=\linewidth]{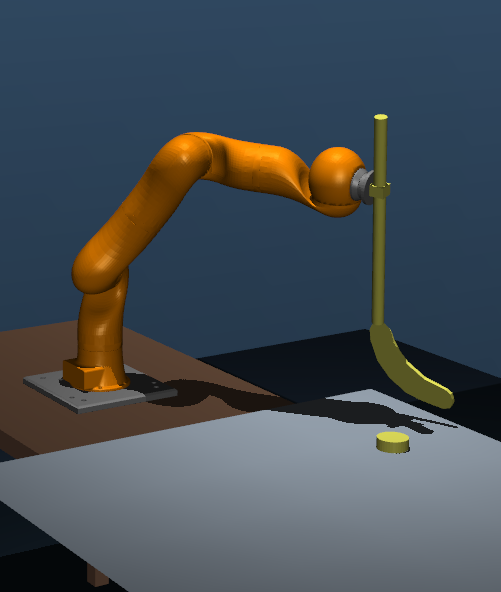}
    \caption{}
    \label{fig:setup_sim}
    \vspace{-4pt}
    \end{subfigure}
    \begin{subfigure}{0.37\linewidth}
    \includegraphics[width=\linewidth]{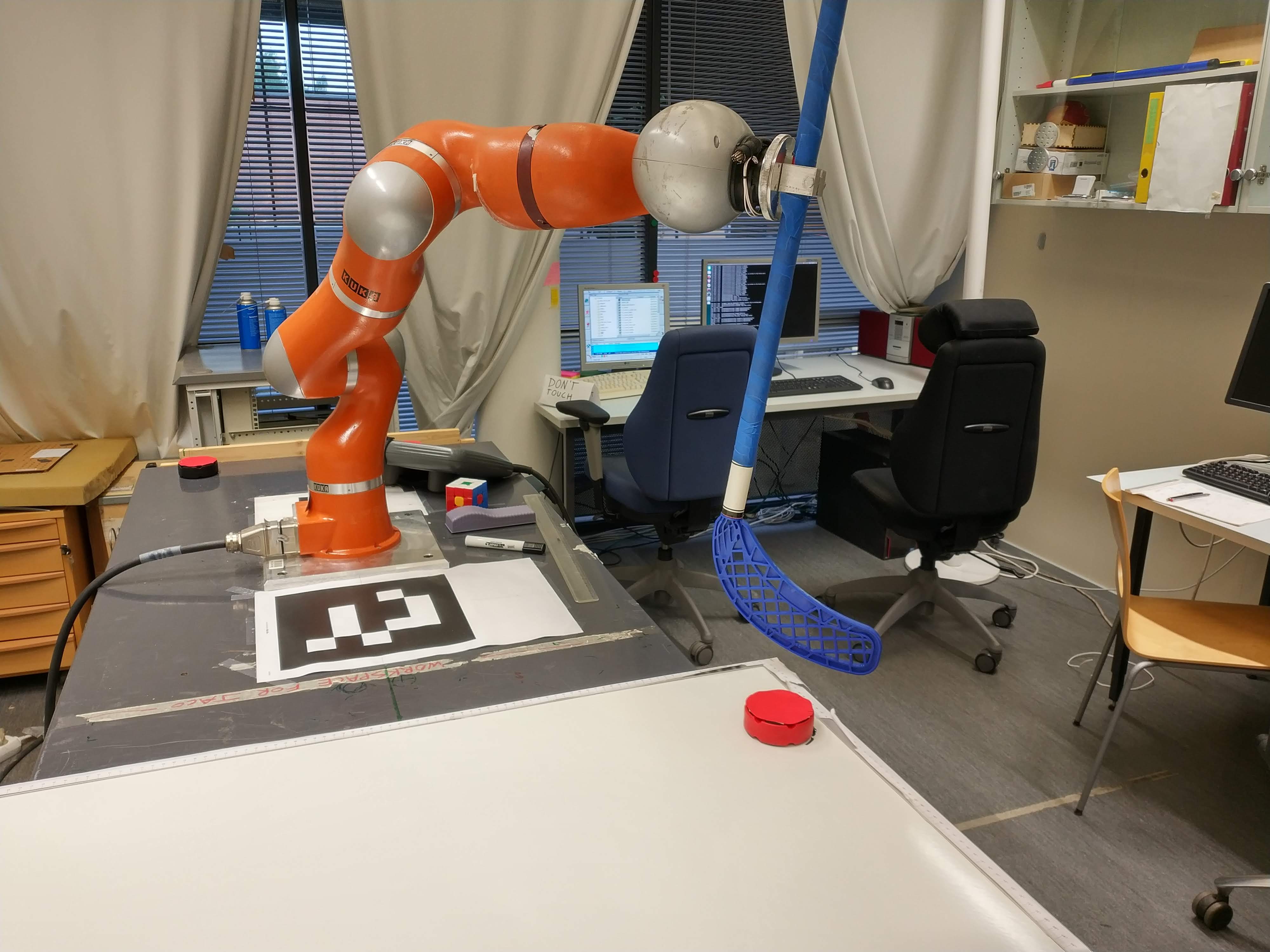}
    \caption{}
    \label{fig:setup_robot}
    \vspace{-4pt}
    \end{subfigure}
    \begin{subfigure}{0.37\linewidth}
    \includegraphics[width=\linewidth]{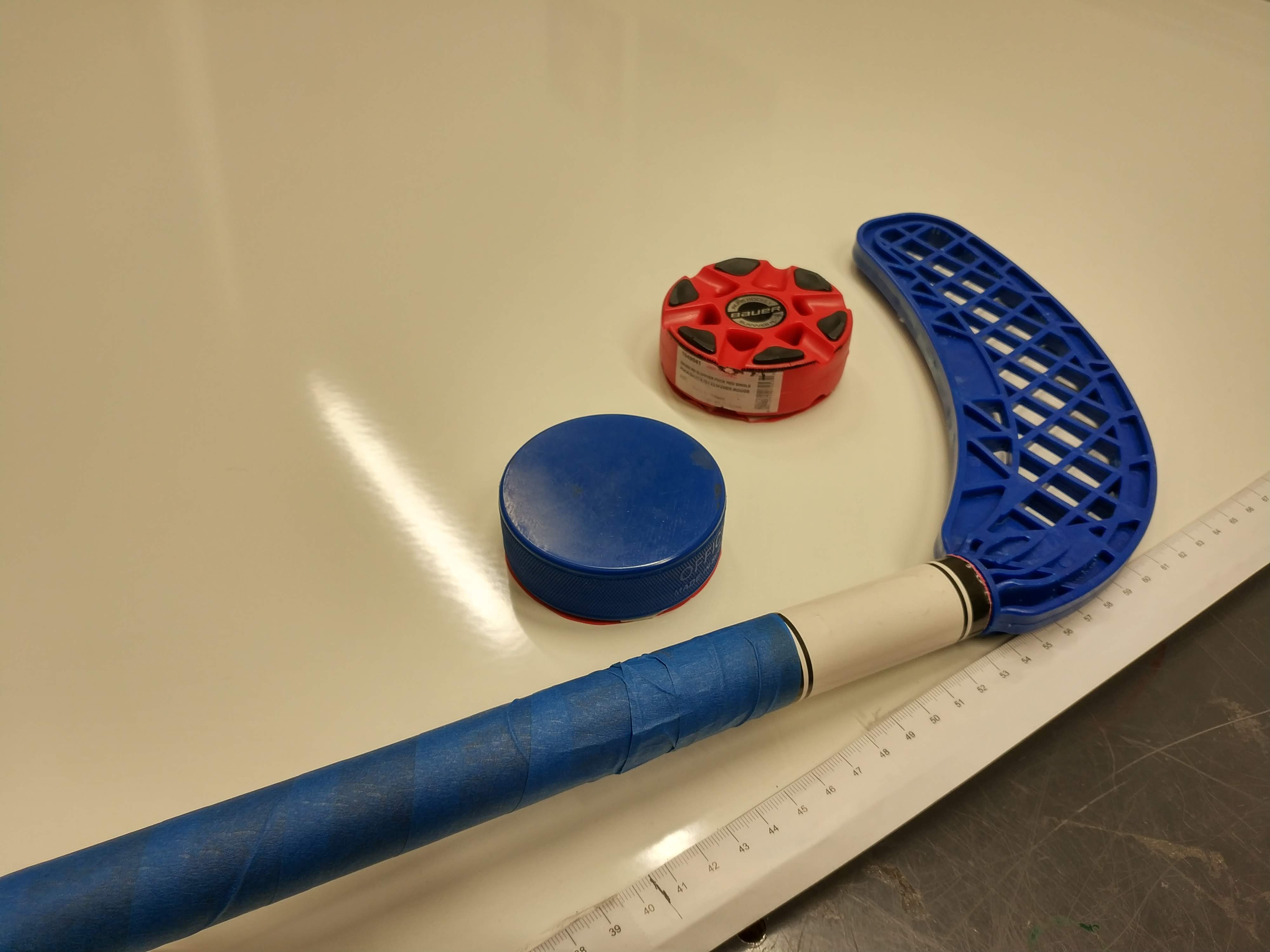}
    \caption{}
    \label{fig:setup_pucks}
    \vspace{-4pt}
    \end{subfigure}
    \caption{The simulated~(\subref{fig:setup_sim}) and real~(\subref{fig:setup_robot}) experimental setups, with a close-up of tools used for the experiments (\subref{fig:setup_pucks})}
    \label{fig:setup}
    \vspace{-12pt}
\end{figure}

For the sim-to-real experiments, we used the same hockey puck setup as in~\cite{arndt2019meta}.
This setup consists of a KUKA LBR4+ robot arm equipped with a floorball stick, a horizontally placed whiteboard, and a two hockey pucks with different masses and friction characteristics: an inline hockey puck and an ice hockey puck.
The goal is to hit the hockey puck with the stick such that the puck stops at a given target location.
Solving this problem requires the model to learn the friction parameters between the puck and the surface it is sliding on, as, after the puck is hit, friction is the main force causing it to stop.
This setup is very similar to FetchSlide, with the additional challenge posed by some phenomena which occur in the physical world, but are not modelled in the simulator, such as hockey blade bending due to elasticity or the whiteboard surface not being perfectly flat and uniform.
The experimental setup and the tools are presented in Figure~\ref{fig:setup}.

The position of the hockeypuck is measured by a top-mounted Kinect camera.
While the camera itself has quite small inaccuracy, we noticed that the system dynamics are not fully deterministic; executing the same trajectory with the same starting puck position (up to a margin of manual positioning error) may produce different results with the standard deviation in position around 5--10 cm, depending on the trajectory and the puck used.
We use the same trajectory generation approach as in the FetchSlide experiments, with the trajectories being generated in joint space instead of Cartesian space.

\subsubsection{Hockeypuck simulation experiments}
The simulated setup was constructed in MuJoCo~\cite{Todorov2012MuJoCoAP}, based on the physical setup.
In order to generate training data from a wide variety of conditions, we randomized the friction between the hockey puck and the whiteboard surface, as well as the mass of the puck.
To account for slight misalignments between the physical and simulated setups, we additionally randomized the starting position of the puck.
We added random noise in a similar fashion to FetchSlide, with noise variance sampled from $\sigma_s^2 \sim \mathcal{U}(0, 0.5)$.
These randomizations make up for a much wider range of system dynamics than tested in FetchSlide, and match the configurations presented in~\cite{arndt2019meta}.

We also compare the performance to model adaptation with MAML~\cite{finn2017maml}.
Due to the gradient update rule in MAML being batched, as opposed to recurrent architectures, we test MAML models trained with different amounts of environment rollouts used per update step, where smaller batch sizes allow the model to adapt with fewer samples, but the updates are less accurate.
The network used for MAML consists of 4 fully connected layers with 128 neurons each.
We performed up to 3 adaptation steps with MAML during training, as more steps led to instability during training and adaptation.
The gradient update step size $\alpha$ is learned during training.

\begin{table}[]
    \centering
    \begin{tabular}{c|c}
        Parameter & Distribution \\
        \hline
        Puck mass & $\mathcal{U}(0.01, 0.1)$ \\
        Linear friction $\mu_x$ & $\mathcal{U}(0.15, 0.95)$ \\
        Linear friction $\mu_y$ & $\mathcal{U}(0.7\mu_x, 1.3\mu_x)$ \\
        Torsional friction & $\mathcal{U}(0.001, 0.05)$ \\
        Rotational friction & $\mathcal{U}(0.01, 0.3)$ \\
        Initial position error & $\mathcal{N}(0, 0.02)$ \\
    \end{tabular}
    \caption{Randomization parameters}
    \label{tab:hp_params}
    \vspace{-12pt}
\end{table}

\begin{figure}
    \centering
    \begin{subfigure}{.35\linewidth}
        \includegraphics[width=\linewidth]{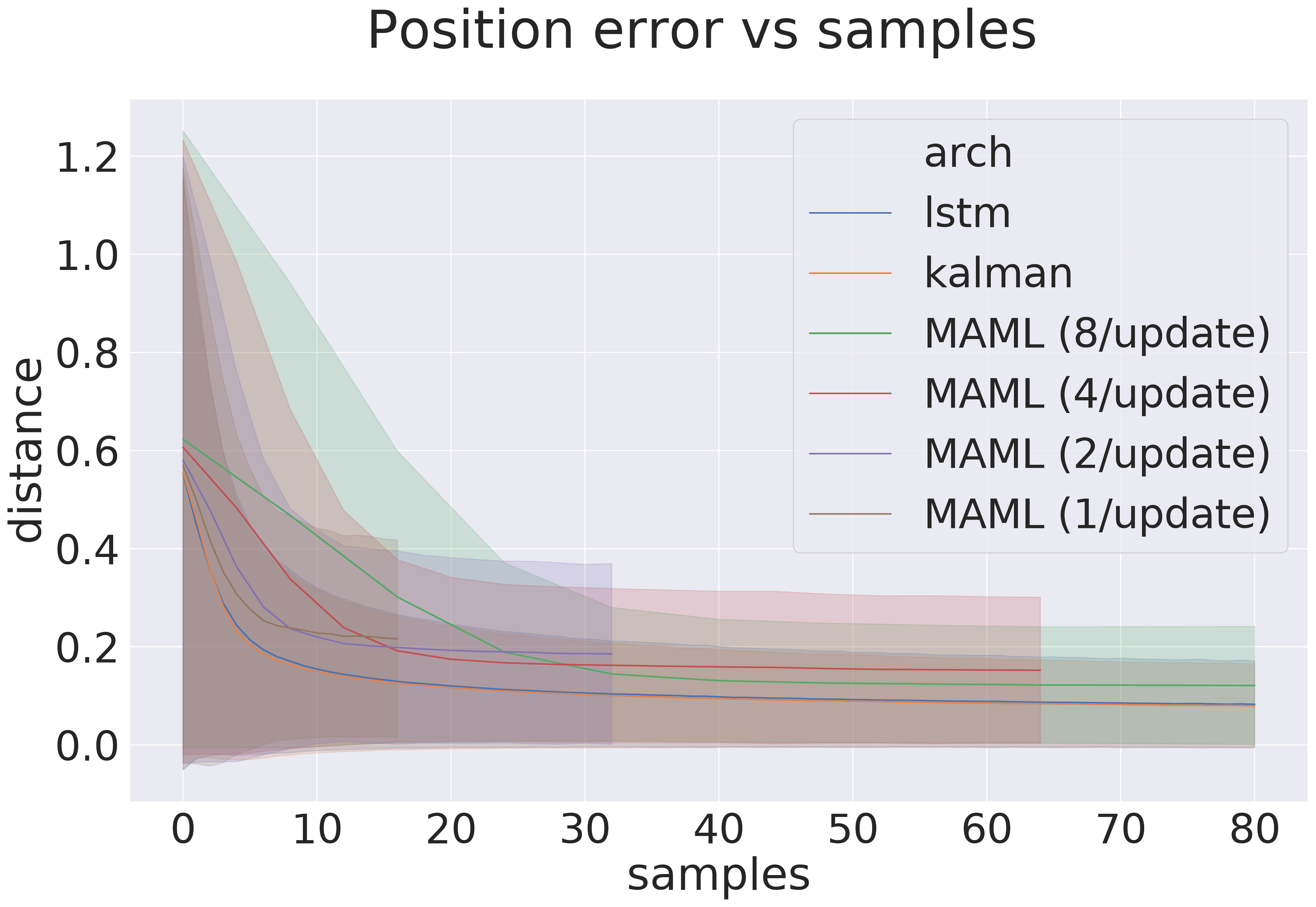}
        \caption{}
        \label{fig:err_method_sim}
    \end{subfigure}
    \begin{subfigure}{.6\linewidth}
        \includegraphics[width=\linewidth]{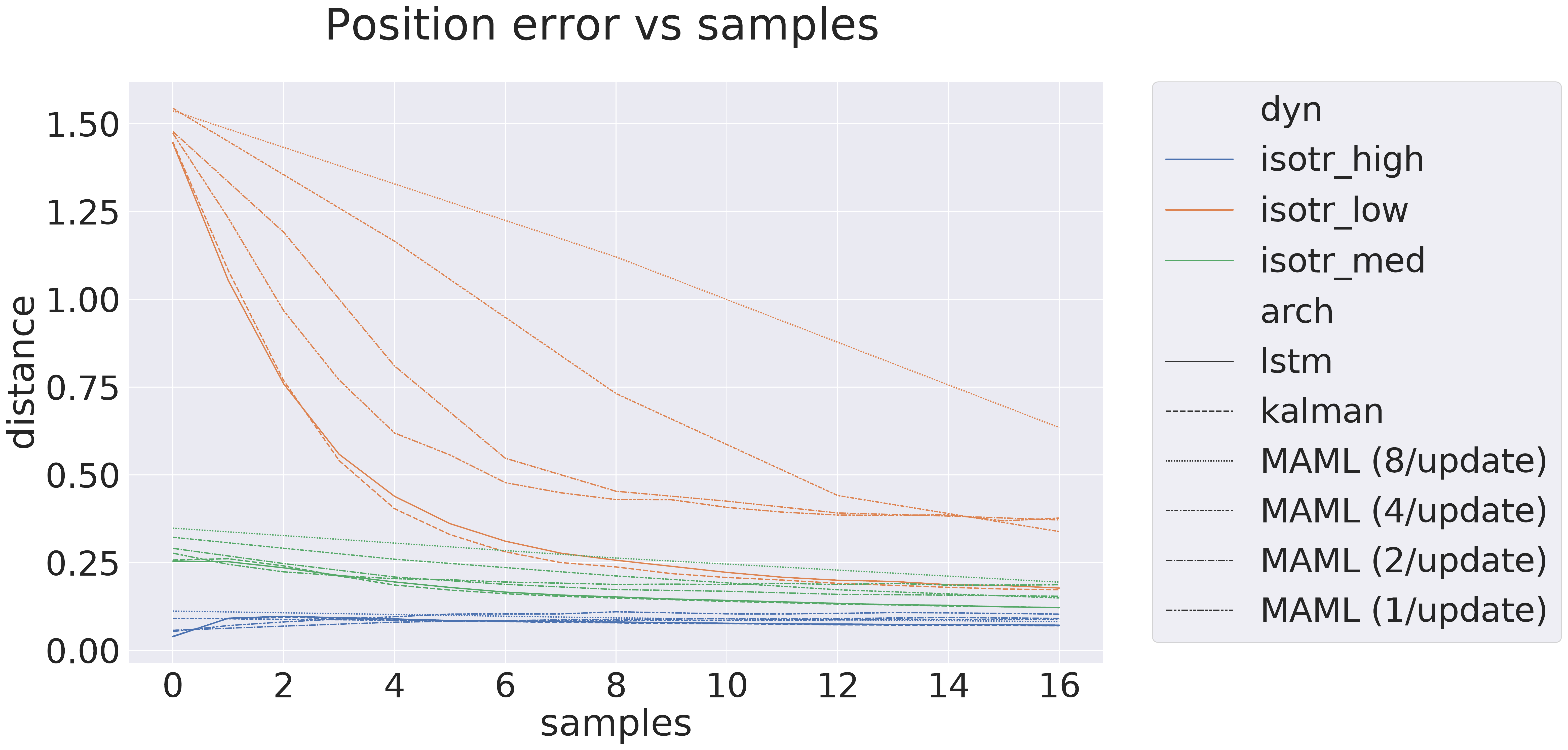}
        \caption{}
        \label{fig:err_per_cond_sim_ls_maml}
    \end{subfigure}
    \begin{subfigure}{\linewidth}
        \includegraphics[width=\linewidth]{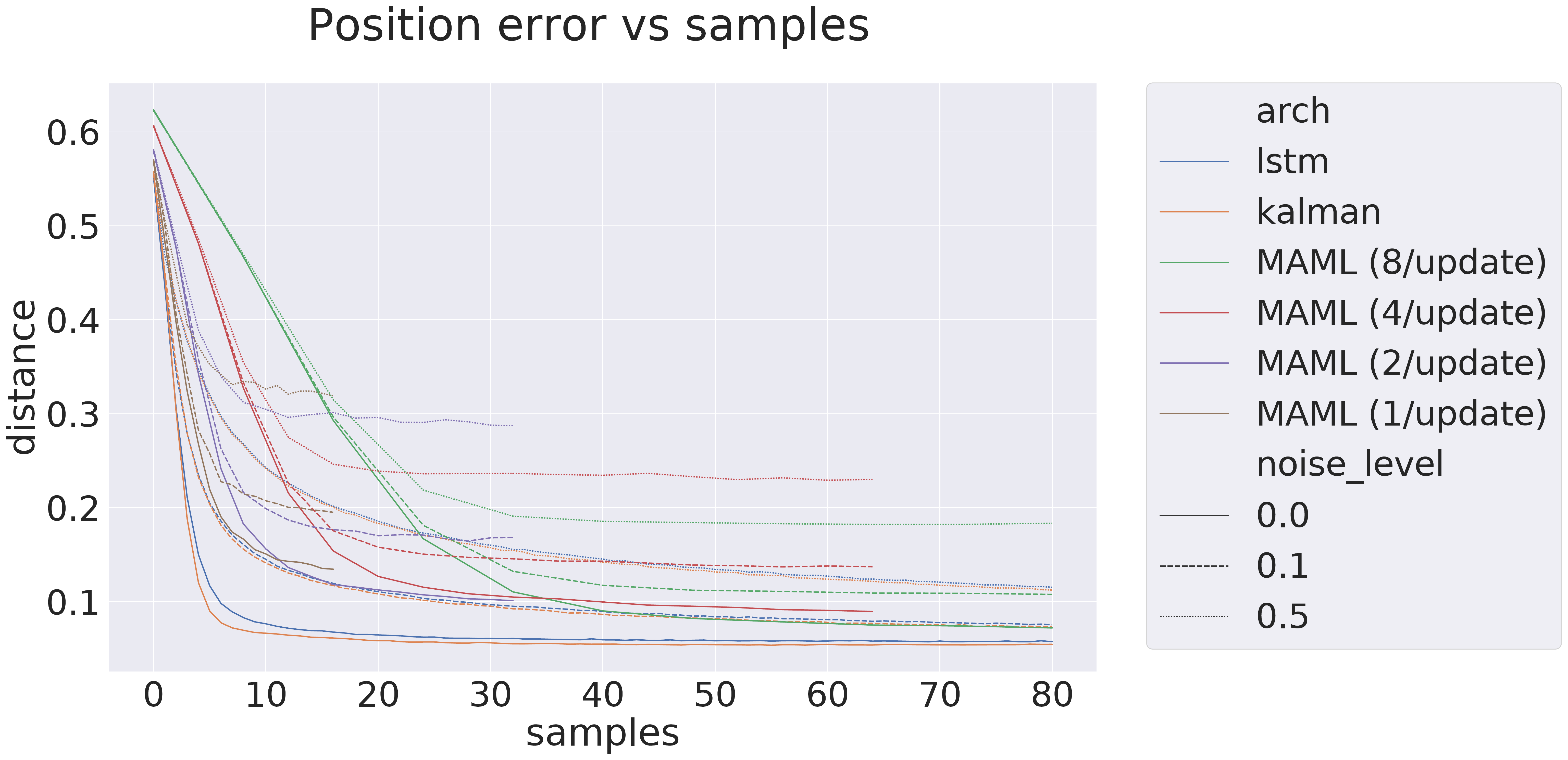}
        \caption{}
        \label{fig:err_dynnoci_sim}
    \end{subfigure}
    \vspace{-2pt}
    \caption{Performance in HockeyPuck (simulation), averaged over all test conditions~(\subref{fig:err_method_sim}), shown separately for different simulated conditions~(\subref{fig:err_per_cond_sim_ls_maml}), and for each noise level (\subref{fig:err_dynnoci_sim})}
    \label{fig:err_hp_sim}
    \vspace{-18pt}
\end{figure}

The results of the evaluation in simulation, as expressed by distance between the predicted and true position in meters, are shown in Figure~\ref{fig:err_hp_sim}.
We can see that, in a scenario where both models were trained and evaluated on data from simulation (which falls within the task distribution seen during training), there is no significant change in performance between the proposed architecture and LSTM; both methods perform the same with similar error margins (the shaded areas indicate the standard deviation of prediction errors).

On this benchmark, both recurrent methods also outperform MAML~\cite{finn2017maml}, as shown in Figure~\ref{fig:err_hp_sim}.
In Figure~\ref{fig:err_method_sim}, we observe that performing MAML updates with 1 sample per update achieves comparable performance improvements at first, but due to noise converges to a large average error of over 20~cm.
Performing more "stable" MAML updates with 8 samples per update achieves comparable, but slightly worse, final performance, yet performs much worse when only a few samples are available.

In Figure~\ref{fig:err_per_cond_sim_ls_maml}, we detail the adaptation error for a selection of different conditions: low, medium and high isotropic frictions in the low sample regime.
We can observe that the adaptation error is the highest in the low friction case.
Figure~\ref{fig:err_dynnoci_sim} shows the error for each method in various noise conditions.
We again observe that both the LSTM and Kalman methods outperform MAML, with the Kalman approach providing slightly better performance.
We also observe that, as expected, the difference between noisy and noiseless conditions for MAML is more pronounced when the update batch size is small.

\subsubsection{Physical experiments}
\begin{figure}
    \centering
    \begin{subfigure}{.49\linewidth}
        \includegraphics[width=\linewidth]{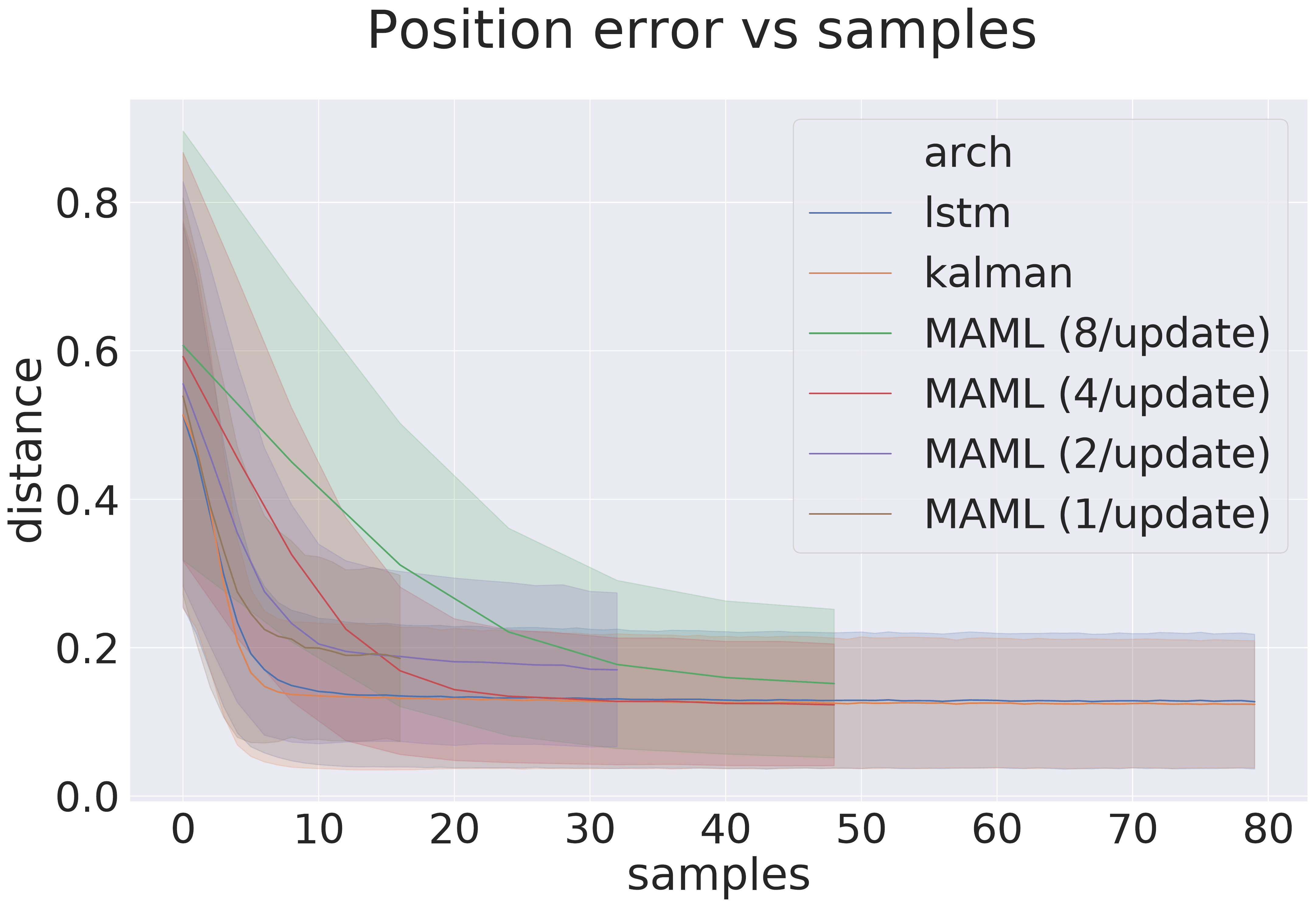}
        \caption{}
        \label{fig:err_method_real}
    \vspace{-2pt}
    \end{subfigure}
    \begin{subfigure}{.49\linewidth}
        \includegraphics[width=\linewidth]{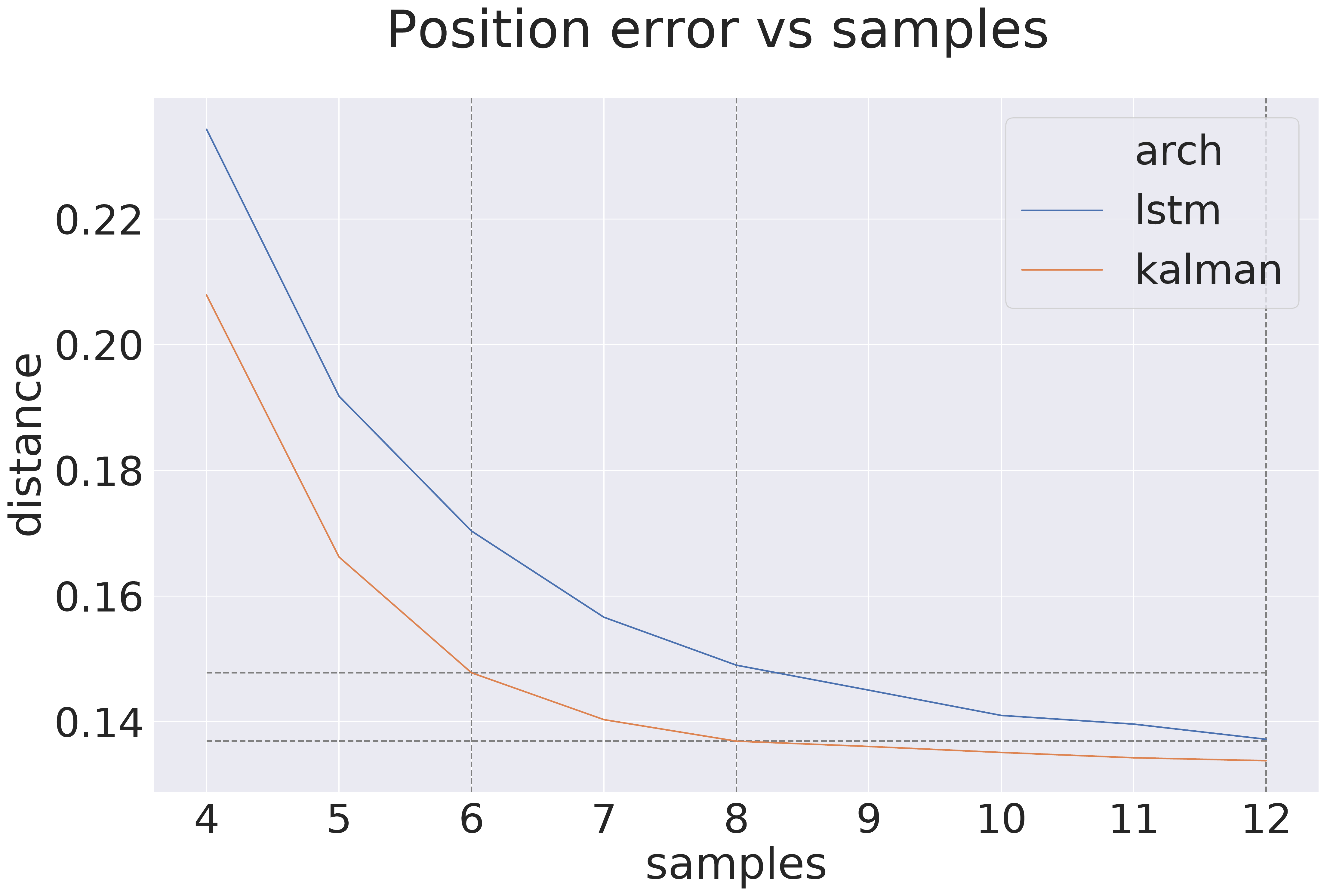}
        \caption{}
        \label{fig:err_zoom_real}
    \vspace{-2pt}
    \end{subfigure}
    \begin{subfigure}{\linewidth}
    \centering
        \includegraphics[width=.8\linewidth]{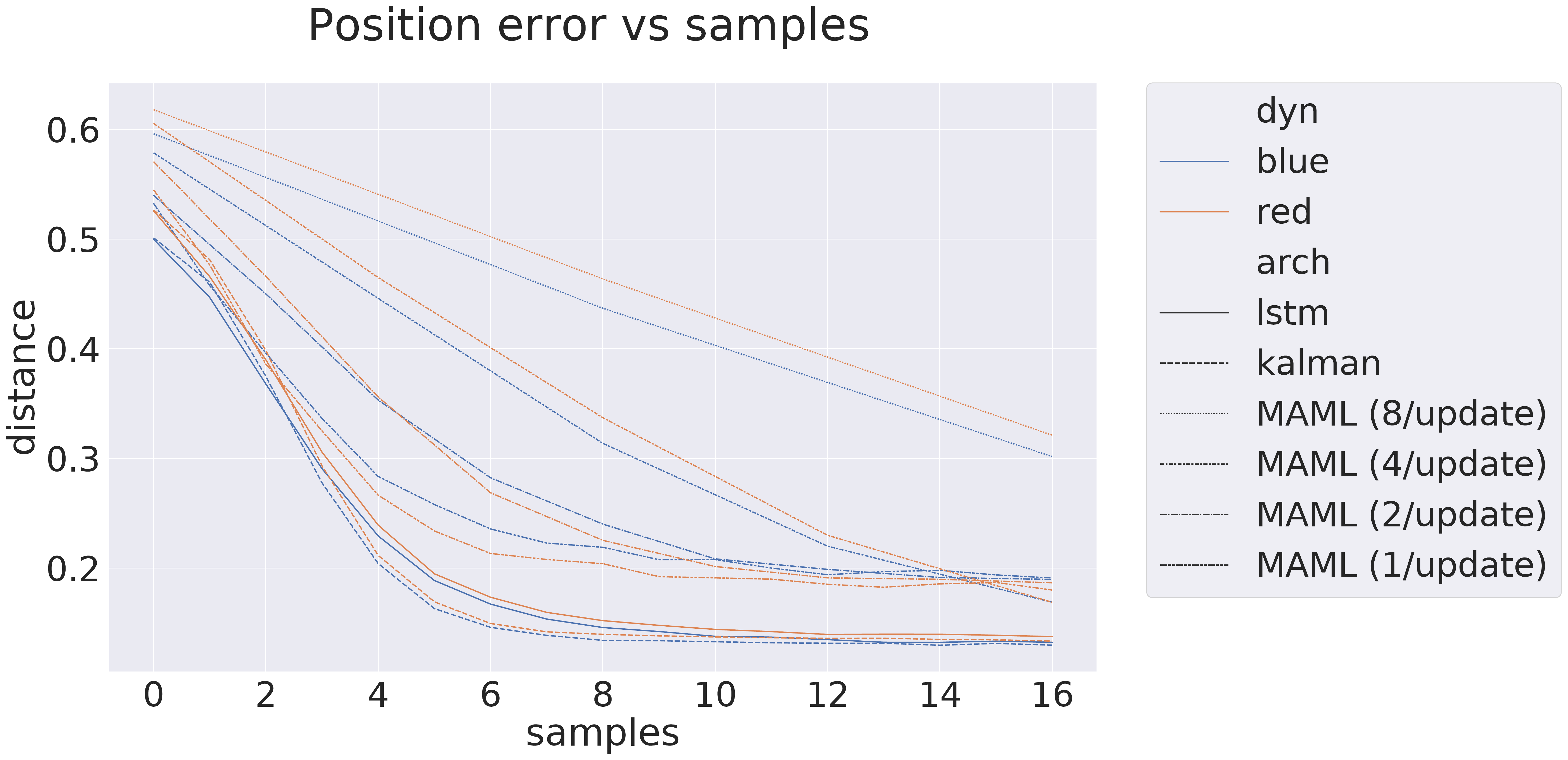}
        \caption{}
        \label{fig:err_dynnoci_real}
    \vspace{-4pt}
    \end{subfigure}
    \caption{Performance in real conditions averaged over two pucks~(\subref{fig:err_method_real}), zoomed in~(\subref{fig:err_zoom_real}), and detailed for each puck in~(\subref{fig:err_dynnoci_real})}
    \label{fig:err_hp_real}
    \vspace{-18pt}
\end{figure}

In order to test the suitability of our method for sim-to-real transfer, we generated random trajectories using the generative model, executed them on the physical setup, recorded the resulting puck positions, and evaluated each method on the resulting data.
The results  are shown in Figure~\ref{fig:err_hp_real}.

In Figure~\ref{fig:err_method_real}, we observe that both recurrent methods outperform MAML in the low sample area.
Analyzing the performance of various model-based MAML updates, we observe that, similarly to simulation, updating with one sample achieves the best performance at first, but converges to a fairly large error.
With 4 samples per update, MAML is able to match the performance of recurrent after about 20 samples (around 3 times more).
We can also observe that the Kalman filter-based method outperforms the blackbox LSTM architecture in the low-sample regime, as highlighted in Figure~\ref{fig:err_zoom_real}.
We observe that the Kalman filter-based approach is able reach error below 14cm with 8 samples, while the LSTM approach requires 12 (50\% more).
We believe that, due to inaccuracies between simulated and real conditions, the real-world data lies a bit out of the training distribution of the model, especially in terms of the noise distribution; thus, having the inference procedure embedded in the graph, can improve generalization to out-of-task distributions. 
In~\cite{finn2018universality}, it was observed that enforcing the inner-loop update to gradient descent improves generalization to classification tasks with data coming from outside the training domain, in comparison to learned update rules; this is because, even for out-of-domain data, gradient descent still constitutes a sensible update rule.
In a similar fashion, based on our results, we can state that the Kalman filter update rules provide better handling of out-of-domain data distributions in comparison to learned update rules, as is the case with LSTM.

In Figure~\ref{fig:err_dynnoci_real}, we show the estimation error for each method for the two hockeypucks we used for the evaluation, zoomed in to the low sample area.
We observe that both recurrent methods outperform MAML, with the proposed Kalman filter-based approach outperforming LSTM for both pucks.

We also compare our results to the results obtained in~\cite{arndt2019meta}, where a reinforcement learning policy was trained in simulation for the same setup and task.
In that work, the average position error for the red puck was 14.4 cm after observing 16 real-world samples; with the model-based adaptation approach, we achieve a comparable error (14.2 cm) after observing only 7 samples.
Similarly, for the blue puck the average error after observing 16 samples was 27.7 cm, while the proposed method achieves the same performance with only 3 samples, reaching the average prediction error of 13.3cm with 16 samples.
After 64 observations, the average error in~\cite{arndt2019meta} was 13.8 cm; with the proposed method, we were able to reach this value with only 7 samples.
Thus, our trained model could be used to find a much better policy (e.g. by backpropagating through the learned environment model or by training an inverse model) than the policy trained in~\cite{arndt2019meta}.

\begin{figure}
    \centering
    \begin{subfigure}{.49\linewidth}
    \includegraphics[width=\linewidth]{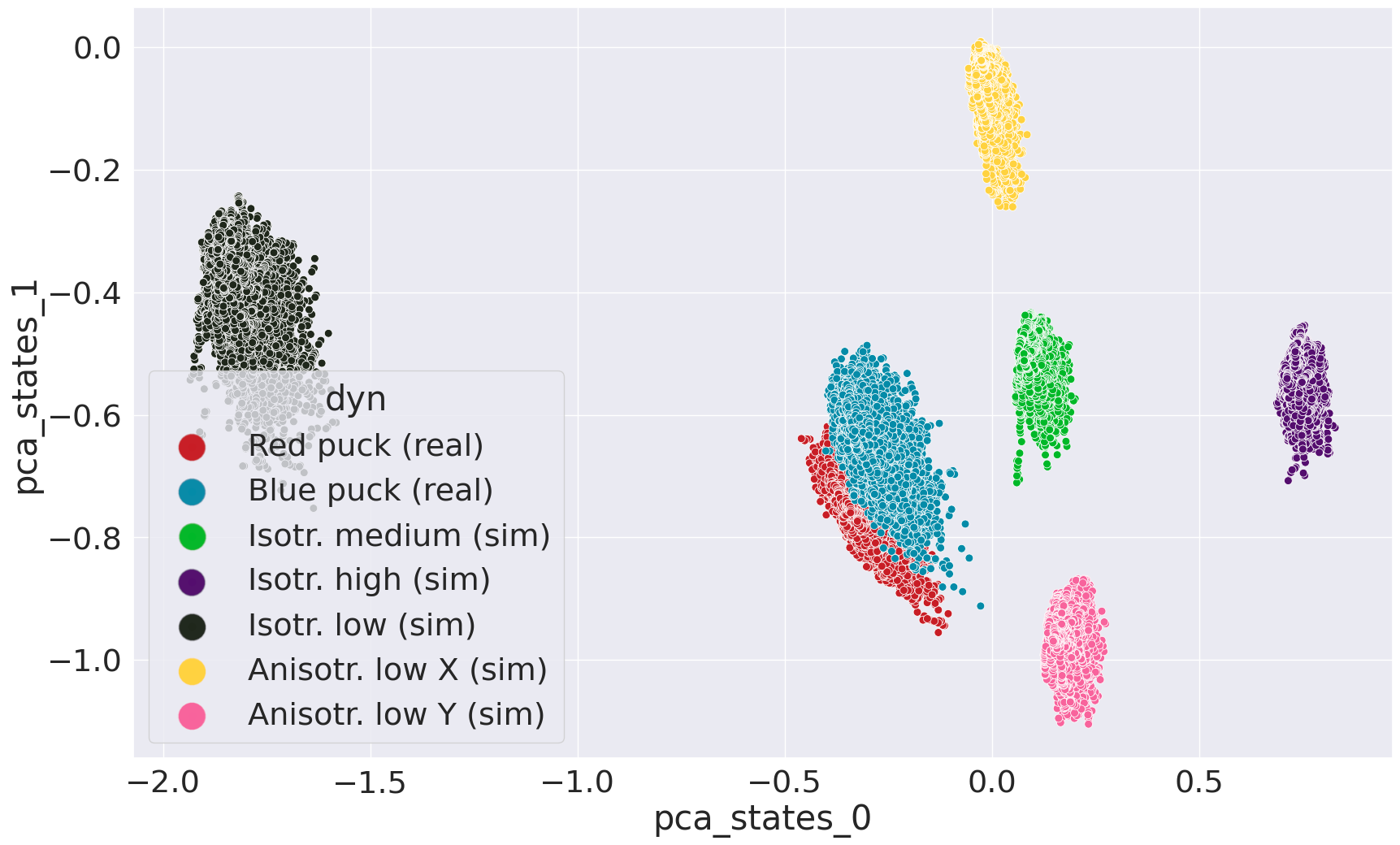}
    \caption{}
    \label{fig:hp_pca_kal_dyn}
    \vspace{-4pt}
    \end{subfigure}
    \begin{subfigure}{.49\linewidth}
    \includegraphics[width=\linewidth]{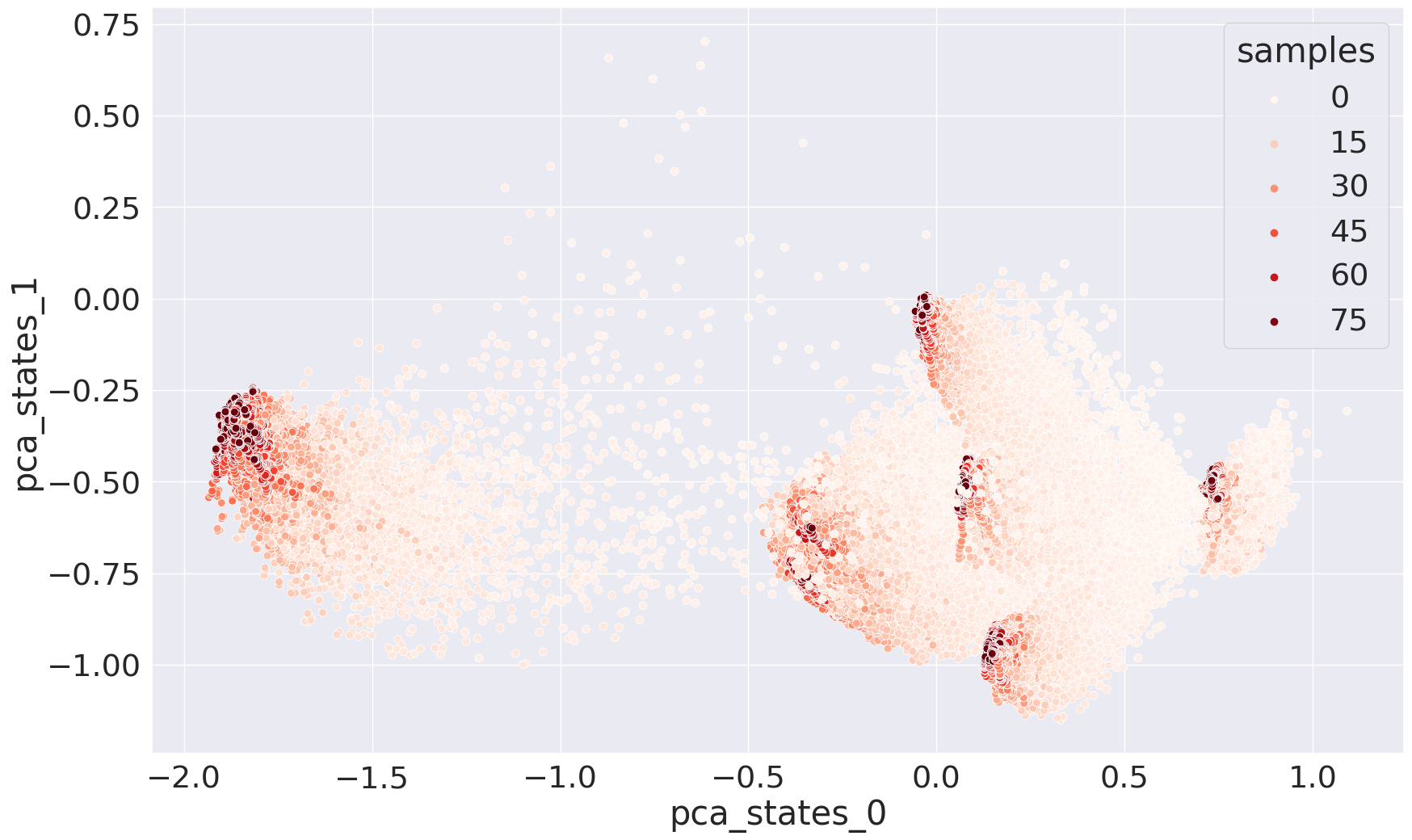}
    \caption{}
    \label{fig:hp_pca_kal_samples}
    \vspace{-4pt}
    \end{subfigure}
    \caption{PCA analysis of hidden states in Hockeypuck for Kalman filter-based adaptation, showing real and simulated domains. (\subref{fig:hp_pca_kal_dyn}) shows clusters after observing 20 samples.}
    \label{fig:hp_pca}
    \vspace{-12pt}
\end{figure}

In Figure~\ref{fig:hp_pca}, we visualize the state space of the Kalman filter using PCA, similarly to previous FetchSlide analysis.
We can also see that the real conditions we used for testing actually lie very close to each other, between medium and low friction, in the direction of anisotropic friction with lower value along $y$.
Based on this, we can say that the randomization range used in the simulator for generating the training data was excessive.
We also suspect that the shift towards anisotropic effects is a result of unmodeled phenomena, such as the hockey blade slightly bending during the hit.
This analysis can be used to search for a range of randomization parameters that encapsulate the real conditions with a much smaller margin of error (by adjusting the parameters given in Table~\ref{tab:hp_params})
Similarly, comparing cluster sizes between real and simulated conditions with varying amounts of noise can provide a more accurate measure about the noise level in the real domain.
Such an analysis would provide a more accurate state prior for the dynamics model and could be used to train a model which achieves even smaller error in the low sample regime with the same training method.

We also observe that the hidden state space has learned friction representations which disentangle the magnitude and direction of friction---the low friction domain lies to the left, with the friction increasing along the first PC.
Based on the position of the anisotropic friction domains, we can state that the second PC encodes the direction of friction---the domain with lower friction along $x$ lies above the isotropic medium friction domain, while the domain with lower $y$ friction lies below it.
Thus, we observe that the method learns interpretable and useful dynamics representations.

\section{Conclusions}
In this paper, we have have presented a novel, uncertainty-aware meta learning algorithm and demonstrated that its performance on a model adaptation in the context of sim-to-real transfer.
The proposed method, which directly takes into account uncertainty, performs better than both an LSTM-based approach (where the optimal way of handling uncertainty is learned) and than MAML (where the uncertainty information is discarded, and only the mean is taken into account).
We also show that model adaptation, as opposed to policy adaptation, is a more data efficient approach---compared to a previous approach~\cite{arndt2019meta}, all model-based methods offer superior performance.
We also demonstrated how the proposed method learns an interpretable, disentangled space of dynamics representations.

We hypothesize that the additional benefit of using a Kalman filter-based approach lies in the way the uncertainty is explicitly handled in the inner loop update;
the method used in the Kalman filter (which is Bayes-optimal for Gaussian noise) generalizes to non-Gaussian distributions better than the blackbox approach learned by the LSTM.

In our experiments, we assumed that the whole trajectory can be generated in advance and passed to the robot for execution.
For some complex problems, however, it is necessary to use  a feedback policy, which adds time complexity to the system.
While the general idea behind our method would also be applicable to such problems, we leave the performance to be evaluated in the future.

Additionally, removing time complexity and using generative trajectory models gives us a natural way of exploring the environment through sampling from the latent distribution of trajectories.
While we showed that this simple method can be sufficient, it is likely that a more informed search strategy (for example based on uncertainty information) would provide more informative samples, and thus adapt to real conditions with fewer samples.
Finding proper search strategies is especially crucial in order to generalize the method for feedback policies, where there is no latent space of safe trajectories to sample from.

It is also interesting to see how uncertainty awareness translates into more general few-shot function estimation, e.g., in classification tasks involving label ambiguity.

\bibliographystyle{ieeetr}
\bibliography{main}  

\begin{thebibliography}{10}

\bibitem{hamalainen2019affordance}
A.~H{\"a}m{\"a}l{\"a}inen, K.~Arndt, A.~Ghadirzadeh, and V.~Kyrki, ``Affordance
  learning for end-to-end visuomotor robot control,'' in {\em IROS}, 2019.

\bibitem{Tobin2017domain}
J.~Tobin, R.~Fong, A.~Ray, J.~Schneider, W.~Zaremba, and P.~Abbeel, ``Domain
  randomization for transferring deep neural networks from simulation to the
  real world,'' in {\em IROS}, 2017.

\bibitem{andrychowicz19learning}
OpenAI, ``Learning dexterous in-hand manipulation,'' {\em IJRR}, vol.~39, 2020.

\bibitem{transferMurtaza}
M.~{Hazara} and V.~{Kyrki}, ``Transferring generalizable motor primitives from
  simulation to real world,'' {\em IEEE RA-L}, vol.~4, 2019.

\bibitem{arndt2019meta}
K.~Arndt, M.~Hazara, A.~Ghadirzadeh, and V.~Kyrki, ``Meta reinforcement
  learning for sim-to-real domain adaptation,'' in {\em ICRA}, 2019.

\bibitem{Nagabandi2019learning}
A.~Nagabandi, I.~Clavera, S.~Liu, R.~S. Fearing, P.~Abbeel, S.~Levine, and
  C.~Finn, ``Learning to adapt in dynamic, real-world environments through
  meta-reinforcement learning,'' in {\em ICLR}, 2019.

\bibitem{schmidhuber:1987:srl}
J.~Schmidhuber, ``Evolutionary principles in self-referential learning. on
  learning now to learn: The meta-meta-meta...-hook,'' diploma thesis,
  Technische Universitat Munchen, Germany, 1987.

\bibitem{finn2017maml}
C.~Finn, P.~Abbeel, and S.~Levine, ``Model-agnostic meta-learning for fast
  adaptation of deep networks,'' in {\em ICML}, 2017.

\bibitem{8461241}
M.~{Hazara} and V.~{Kyrki}, ``Speeding up incremental learning using data
  efficient guided exploration,'' in {\em ICRA}, 2018.

\bibitem{ortega2019metalearning}
P.~A. Ortega~et al., ``Meta-learning of sequential strategies,'' tech. rep.,
  DeepMind, 2019.

\bibitem{kalman1960kalman}
R.~E. Kalman, ``A new approach to linear filtering and prediction problems,''
  {\em Transactions of the ASME--Journal of Basic Engineering}, 1960.

\bibitem{haarnoja2016backprop}
T.~Haarnoja, A.~Ajay, S.~Levine, and P.~Abbeel, ``Backprop kf: Learning
  discriminative deterministic state estimators,'' in {\em NIPS}, 2016.

\bibitem{becker2019recurrent}
P.~Becker, H.~Pandya, G.~Gebhardt, C.~Zhao, C.~Taylor, and G.~Neumann,
  ``Recurrent kalman networks: factorized inference in high-dimensional deep
  feature spaces,'' in {\em ICML}, 2019.

\bibitem{hochreiter2001learning}
S.~Hochreiter, A.~S. Younger, and P.~R. Conwell, ``Learning to learn using
  gradient descent,'' in {\em ICANN}, 2001.

\bibitem{flennerhag2019metalearning}
S.~Flennerhag, A.~A. Rusu, R.~Pascanu, H.~Yin, and R.~Hadsell, ``Meta-learning
  with warped gradient descent,'' in {\em ICLR}, 2020.

\bibitem{stadie2018emaml}
B.~Stadie, G.~Yang, R.~Houthooft, P.~Chen, Y.~Duan, Y.~Wu, P.~Abbeel, and
  I.~Sutskever, ``The importance of sampling in meta-reinforcement learning,''
  in {\em NIPS}, 2018.

\bibitem{finn2018platipus}
C.~Finn, K.~Xu, and S.~Levine, ``Probabilistic model-agnostic meta-learning,''
  in {\em NIPS}, 2018.

\bibitem{yoon2018bmaml}
J.~Yoon, T.~Kim, O.~Dia, S.~Kim, Y.~Bengio, and S.~Ahn, ``Bayesian
  model-agnostic meta-learning,'' in {\em NIPS}, 2018.

\bibitem{gordon2018metalearning}
J.~Gordon, J.~Bronskill, M.~Bauer, S.~Nowozin, and R.~E. Turner,
  ``Meta-learning probabilistic inference for prediction,'' in {\em ICLR},
  2019.

\bibitem{tan2018sim}
J.~Tan, T.~Zhang, E.~Coumans, A.~Iscen, Y.~Bai, D.~Hafner, S.~Bohez, and
  V.~Vanhoucke, ``Sim-to-real: Learning agile locomotion for quadruped
  robots,'' in {\em RSS}, 2018.

\bibitem{sadeghi17sim2real}
F.~Sadeghi, A.~Toshev, E.~Jang, and S.~Levine, ``Sim2real view invariant visual
  servoing by recurrent control,'' in {\em CVPR}, 2018.

\bibitem{Song2020RapidlyAL}
X.~Song, Y.~Yang, K.~Choromanski, K.~Caluwaerts, W.~Gao, C.~Finn, and J.~Tan,
  ``Rapidly adaptable legged robots via evolutionary meta-learning,'' in {\em
  IROS}, 2020.

\bibitem{gupta18metareinforcement}
A.~Gupta, R.~Mendonca, Y.~Liu, P.~Abbeel, and S.~Levine, ``Meta-reinforcement
  learning of structured exploration strategies,'' in {\em NIPS}, 2018.

\bibitem{rothfuss2018promp}
J.~Rothfuss, D.~Lee, I.~Clavera, T.~Asfour, and P.~Abbeel, ``Promp: Proximal
  meta-policy search,'' in {\em ICLR}, 2019.

\bibitem{clavera2018model}
I.~Clavera, J.~Rothfuss, J.~Schulman, Y.~Fujita, T.~Asfour, and P.~Abbeel,
  ``Model-based reinforcement learning via meta-policy optimization,'' in {\em
  CoRL}, 2018.

\bibitem{openaigym}
G.~Brockman, V.~Cheung, L.~Pettersson, J.~Schneider, J.~Schulman, J.~Tang, and
  W.~Zaremba, ``Openai gym,'' 2016.

\bibitem{ghadirzadeh2017deep}
A.~Ghadirzadeh, A.~Maki, D.~Kragic, and M.~Björkman, ``Deep predictive policy
  training using reinforcement learning,'' in {\em IROS}, 2017.

\bibitem{Kingma14Adam}
D.~P. Kingma and J.~Ba, ``Adam: {A} method for stochastic optimization,'' in
  {\em ICLR}, 2015.

\bibitem{finn2018universality}
C.~Finn and S.~Levine, ``Meta-learning and universality: Deep representations
  and gradient descent can approximate any learning algorithm,'' in {\em ICLR},
  2018.

\bibitem{Todorov2012MuJoCoAP}
E.~Todorov, T.~Erez, and Y.~Tassa, ``Mujoco: A physics engine for model-based
  control,'' in {\em IROS}, 2012.

\end{thebibliography}


\end{document}